\documentclass[sigconf]{acmart}

\AtBeginDocument{%
  \providecommand\BibTeX{{%
    \normalfont B\kern-0.5em{\scshape i\kern-0.25em b}\kern-0.8em\TeX}}}

\usepackage{tikz}
\usepackage{amsmath}
\usepackage{longtable}

\usepackage{stfloats}
\usepackage{filecontents}
\usepackage{enumerate}
\usepackage{bbding}
\usepackage{pifont}
\usepackage{wasysym}

\usepackage{amssymb}
\usepackage{setspace}
\usepackage{balance}
\usepackage{algorithm}
\usepackage{algpseudocode}

\usepackage{rotating}
\usepackage{verbatim}
\usepackage{amsfonts}
\usepackage{amscd}
\usepackage{microtype}
\usepackage{array}
\usepackage{chngpage}
\usepackage{graphicx}
\usepackage{multirow}
\usepackage{caption}
\usepackage{subfigure}
\usepackage{lipsum}
\usepackage{float}
\usepackage{bbding}
\usepackage{pifont}
\usepackage{makecell}
\usepackage{bm}
\usepackage{color,xcolor}
\usepackage{booktabs} 
\usepackage{verbatim}
\usepackage{pifont}
\usepackage{array,multirow}
\usepackage{makecell}
\usepackage{CJK}
\usepackage{appendix}  
\usepackage{float}
\usepackage{colortbl}
\usepackage{cleveref}
\usepackage{fancyhdr}
\usepackage{eqparbox}

\usepackage{filecontents}
\usepackage{color}
\usepackage{listings}

\usepackage{tikz}
\usepackage[utf8]{inputenc} 
\usepackage[T1]{fontenc}
\newcommand\toolName{NeurDP}
\newcommand\lir{LIR}
\newcommand\hir{HIR}
\newcommand\cdg{UDG}
\newcommand\otu{OTU}

\newcommand\tw{TU}

\newcommand{\ignore}[1]{}

\newcommand{\del}[1]{}

\newcommand{\revise}[1]{\textcolor{black}{#1}}

\copyrightyear{2022}
\acmYear{2022}
\setcopyright{rightsretained}
\acmConference[ACSAC '22]{Annual Computer Security Applications Conference}{December 5--9, 2022}{Austin, TX, USA}
\acmBooktitle{Annual Computer Security Applications Conference (ACSAC '22), December 5--9, 2022, Austin, TX, USA}
\acmDOI{10.1145/3564625.3567998}
\acmISBN{978-1-4503-9759-9/22/12}

\begin{document}

\vspace{-15pt}
\title{Boosting Neural Networks to Decompile Optimized Binaries}


\author{Ying Cao, Ruigang Liang}
\authornote{Both authors contributed equally to this research.}
\email{{caoying,liangruigang}@iie.ac.cn}
\affiliation{%
  \institution{SKLOIS, IIE, CAS\footnotemark[3]}
  \institution{School of CyberSecurity, UCAS\footnotemark[4]}
  \city{Beijing}
  \country{China}}

\author{Kai Chen}
\authornote{Corresponding Author.}
\email{chenkai@iie.ac.cn}
\affiliation{%
  \institution{SKLOIS, IIE, CAS\footnotemark[3]}
  \institution{School of CyberSecurity, UCAS\footnotemark[4]}
  \institution{BAAI\footnotemark[5]}
  \city{Beijing}
  \country{China}}

\author{Peiwei Hu}
\email{hupeiwei@iie.ac.cn}
\affiliation{%
  \institution{SKLOIS, IIE, CAS\footnotemark[3]}
  \institution{School of CyberSecurity, UCAS\footnotemark[4]}
  \city{Beijing}
  \country{China}}




\begin{abstract}
Decompilation aims to transform a low-level program language (LPL) (eg., binary file) into its functionally-equivalent high-level program language (HPL) (e.g., C/C++). It is a core technology in software security, especially in vulnerability discovery and malware analysis. 
In recent years, with the successful application of neural machine translation (NMT) models in natural language processing (NLP), researchers have tried to build neural decompilers by borrowing the idea of NMT. 
They formulate the decompilation process \revise{as}\del{into} a translation problem between LPL and HPL, aiming to reduce the human cost required to develop decompilation tools and improve their generalizability.
However, state-of-the-art learning-based decompilers do not cope well with compiler-optimized binaries. Since real-world binaries are mostly compiler-optimized, decompilers that do not consider optimized binaries have limited practical significance.
In this paper, we propose a novel learning-based approach named \textit{\toolName}, that targets compiler-optimized binaries. \textit{\toolName} uses a graph neural network (GNN) model to convert LPL to an intermediate representation (IR), which bridges the gap between source code and optimized binary. We also design an Optimized Translation Unit (OTU) to split functions into smaller code fragments for better translation performance.
Evaluation results \revise{on datasets containing various types of statements} show that \textit{\toolName} can decompile optimized binaries with 45.21\% higher accuracy than state-of-the-art neural decompilation frameworks.
\end{abstract}

\maketitle

\section{Introduction}
\label{sec:intro}


\renewcommand{\thefootnote}{\fnsymbol{footnote}}
\footnotetext[3]{Institute of Information Engineering, Chinese Academy of Sciences.}
\footnotetext[4]{University of Chinese Academy of Sciences}
\footnotetext[5]{Beijing Academy of Artificial Intelligence}

\renewcommand*{\thefootnote}{\arabic{footnote}}

In recent years, deep learning has made remarkable achievements in the fields of code comprehension and reverse engineering. Some work is devoted to using neural networks to learn representations of source code, such as GitHub Copilot~\cite{copilot} and CodeBERT~\cite{feng2020codebert}, which are widely used in automatic code generation/completion and automatic comment generation. Similarly, many previous studies leverage neural networks to learn binary or assembly code representation. They perform well on binary-based downstream analysis tasks, including code clone detection~\cite{ding2019asm2vec}, malicious code detection~\cite{downing_deepreflect_2021}, and disassembly~\cite{pei2020xda}. The application of deep learning in software analysis and software reverse engineering\del{~(SRE)} significantly reduces \del{the}human resources and time costs, no matter from the view of developers or analysts. 
In addition, compared to traditional tools, the faster speed of deep neural-based disassembly approaches~\cite{pei2020xda} makes them a powerful engine for downstream models like malware classification. It is meaningful to study how to make neural network (NN) models work well in \del{SRE}\revise{software reverse engineering} and software analysis. 

Decompilation is a core technology in \del{SRE}\revise{software reverse engineering} and software analysis (e.g., vulnerability discovery~\cite{elsabagh2020firmscope, tian2018attention} and malware analysis~\cite{hernandez2020bigmac, yakdan2016helping}), especially adopted in the analysis of commercial software whose source code is not available. The decompilation process can be defined as translating a low-level PL (LPL) (e.g., binary file) into its functionally-equivalent high-level PL (HPL) (e.g., C/C++). 
Developing a traditional decompiler requires much work to manually reverse multiple binaries to analyze and summarize the heuristic rules used in the decompiler.
The well-known open-source decompiler RetDec~\cite{kvroustek2017retdec} has hundreds of developers who contributed code to it. However, since they released the prototype in 2017, the decompilation performance is still unsatisfactory~\cite{liu2020far}.
Due to the limited number of binaries analyzed by PL experts, the decompilation rules are often incomplete. Moreover, the rules might change with instruction set architecture, compilers, and HPL. It is hard to construct a complete decompilation rule set. Therefore, decompilers built on human-defined rules need to iterate continuously by summarizing the rules and collecting feedback on the errors encountered by users. 
With the help of neural-based approaches, \del{the}expert efforts \del{for generalizing and revising}\revise{to generalize and revise} rules could be largely reduced.

Several neural-based  approaches~\cite{katz2018using,katz2019towards,fu2019coda} are explored to decompile LPL to HPL, hoping that the learning algorithms can automatically learn the mapping rules between LPL and HPL. They all used assembly code (ASM) as LPL and source code or abstract syntax tree (AST) as HPL to train an end-to-end model. These schemes perform various preprocessing on the input and output and design various model architectures according to code characteristics. However, they still suffer \del{from}a severe drawback: \textit{none of these works can appropriately handle the decompilation of optimized code}. 
As compiler optimization is ubiquitously used, the ability to decompile optimized code is essential for the practical application of neural-based approaches.

Through extensive analysis and experimentation, we found that it is not easy to train an end-to-end decompilation model that can handle optimized LPL.
Since deep learning models are data-driven, a high-quality training set is critical to the model's performance. Most models used in the source code or reverse engineering domains rely on large, high-quality supervised or unsupervised datasets. In contrast, there are very few mature datasets in the field of decompilation, and datasets used in previous neural-based decompilation studies are not built for the decompilation of optimized LPL. 
Without a well-labeled dataset, it is difficult for the model to learn the mapping rules between HPL and LPL.
Although many open-source projects exist in the real world, the source code and binary code cannot be completely matched at the statement level due to code optimization (e.g., dead code elimination), making it inaccurate to directly use source code as the labels for the optimized binaries.
We summarize the challenges as follows.

\vspace {1pt}\noindent\textbf{Challenges. C1:} It is well known that statements of HPL are often significantly refactored during compiler optimization, which makes it challenging to make an exact match between the semantics of LPL and HPL.
For example, dead code elimination causes certain statements in HPL not to appear in LPL. Loop unwinding can cause some code to appear multiple times in the binary and \del{appear}only once in the source code. These optimization strategies all lead to the code structure and semantic information in the text level of LPL being quite different from HPL. 
In some cases, textually similar HPL codes (only some variable names differ) can correspond to completely different LPL codes, and vice versa.
Therefore, it is not feasible to directly train end-to-end models using HPL as the label of LPL, which makes it challenging to capture the decompilation rules.

\vspace {1pt}\noindent\textbf{C2:} 
Splitting LPL and HPL into code fragments with correct correspondence is a nontrivial task. Previous work typically utilizes functions or basic blocks (BBs) as input units for training neural models.
However, the number of instructions in a function or a BB can be infinite (up to 1,000 instructions), which is hard to handle appropriately by neural network models. 
Therefore, it is essential to split the BB into finer-grained units, which can effectively reduce the model's difficulty in learning the decompiled rules.
One straightforward method is to split the LPL or HPL based on debug information. However, the LPL and HPL mapped by the debug information are inaccurate, especially for optimized binaries.
For example, the dead code in the HPL will also be mapped onto the LPL along with the live code.
Another straightforward way is to set a maximum fragment length and split the basic block into fragments. However, several statements in a fragment may have over one independent feature. For example, there are three independent features (data flows) in the code segment ``$a = 0; b = c + d; call(c);$''. Thus it is difficult for the model to properly encode it into a single vector representing its function or semantics. 
Splitting data \revise{dependency}\del{flow} graphs \revise{(DDG)}\del{DFG} may solve this problem, but it is also a difficult task. Worse still, the \del{DFGs}\revise{DDGs} of LPL and HPL are quite different because of compiler optimization.

\vspace {1pt}\noindent\textbf{Our approach}.
In this paper, we propose an NN-based decompilation framework called \text{\toolName}\footnote{\text{\toolName} (Neural Decompilation)}. \textit{\toolName} uses a neural network model to translate LPL into an optimized IR (IR decompiler) to address C1 instead of directly translating LPL into HPL. As we know, the compiler first generates the intermediate representation (IR) code during the compilation process and performs most of the optimization strategies on the IR. Therefore, the structural differences between the optimized IR and LPL are much more minor than those between HPL and LPL.
Compared to previous end-to-end neural decompilers, \textit{\toolName} can cope with the decompilation problem of compiler-optimized LPL. Finally, \textit{\toolName} converts IR statement to HPL statement directly.  

Specifically, to train a well-performing IR decompiler model, we design a splitting technique called Optimal Translation Unit (\otu) to address C2. \textit{\otu} splits BBs into smaller pairs of LPL and \textit{\hir} fragments. The statements in each fragment have data dependencies and can be synthesized into one feature. \textit{\otu} helps build a high-quality training set for our NN model.

To evaluate the accuracy of \textit{\toolName}, we use several programs randomly generated using our tool which is developed by cfile~\cite{cfile} and regular expressions, including 500 lines of code. 
Since our goal is to boost the neural network's ability for decompilation, we compare \textit{\toolName} with related studies using neural networks (e.g., Coda~\cite{fu2019coda} and Neutron~\cite{liang2021neutron}).
Experimental results show that \textit{\toolName} is 5.8\%-27.8\% more accurate than Coda on unoptimized code. \del{What is more}\revise{Moreover}, \textit{\toolName} can handle compiler-optimized code well, while Coda is incapable of action.
\textit{\toolName} can decompile optimized binaries with 45.21\% higher accuracy than another neural decompilation Neutron~\cite{liang2021neutron}.
According to our evaluation, the introduction of \textit{\otu} and IR mechanisms in our model improves the accuracy by 4.1\%-71.23\% compared to using the model directly in the optimized code.


\vspace {2pt}\noindent\textbf{Contributions.} Our main contributions are outlined below: 

\noindent $\bullet$ We design a novel neural machine decompilation technique. It is the first neural-based decompiler that can handle compiler-optimized code.

\noindent $\bullet$ We design an optimal translation unit (OTU) scheme, which can help other researchers form a sound dataset for training the IR decompilation model or other applications.

\noindent $\bullet$ We implement our techniques and conduct extensive evaluations. 
The results show that \textit{\toolName} is much better than the state-of-the-art neural-based decompilers, especially for optimized code. \revise{We release our dataset and the NN parameters on GitHub\footnote{https://github.com/zijiancogito/neur-dp-data.git}}.


\begin{figure*}[!ht]
\centering
\includegraphics[width=0.85\textwidth]{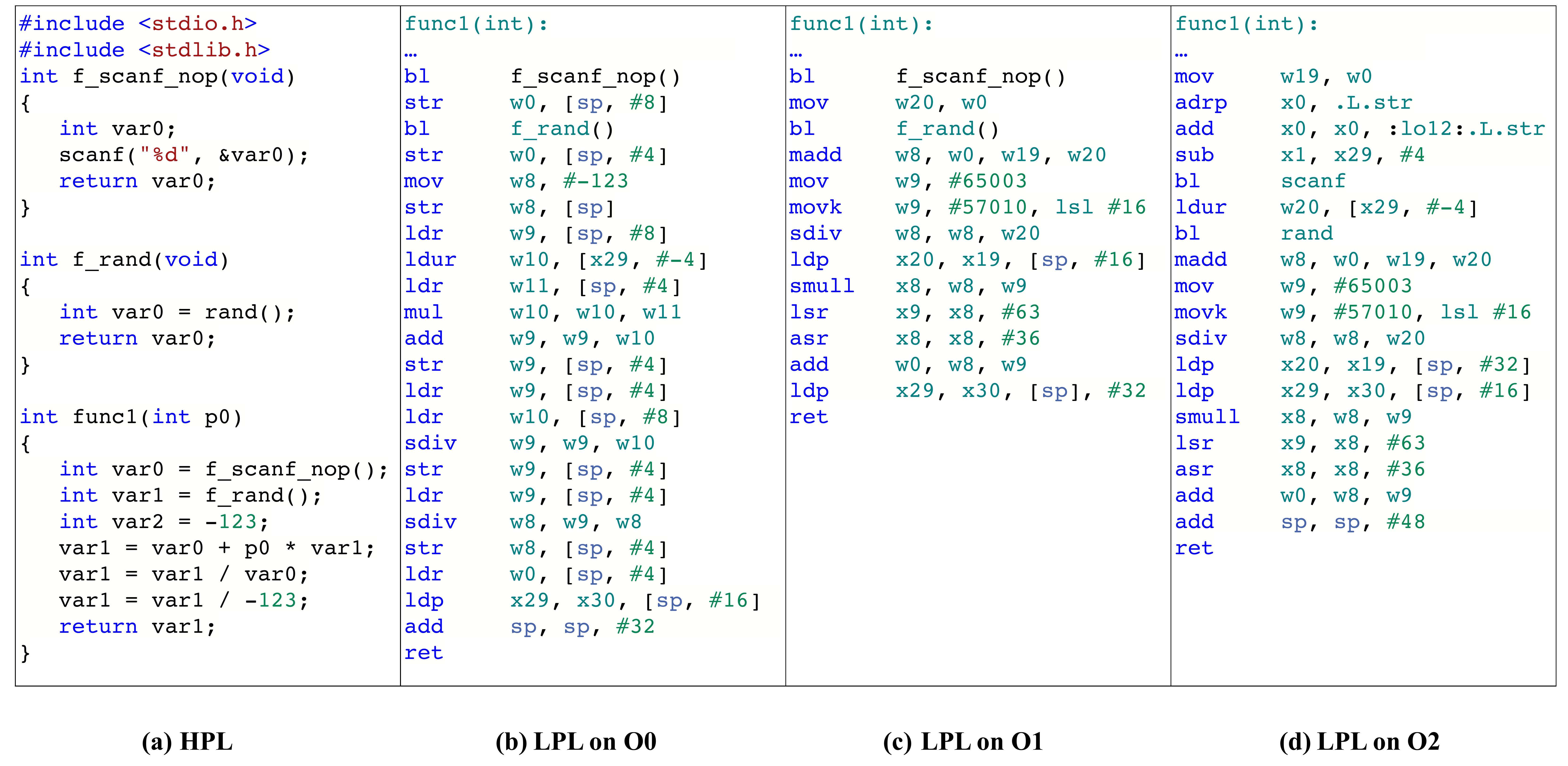}
\vspace{-5pt}
\caption{An example showing different compiler optimization levels}
\label{o03example}
\vspace{-8pt}
\end{figure*}


\section{Background}
\label{sec:background}


\subsection{Compilation and Optimization} 

Compilation translates HPL (e.g., C/C++) into LPL (e.g., machine code) that can be run on the target CPU (e.g., X86, ARM). Due to the differences between the two PLs, unnecessary information (e.g., symbols) for the CPU is usually removed. 
Also, in this process, optimization technologies are designed to minimize or maximize some attributes of the executable program, e.g., to reduce the program’s memory usage. Note that the execution results of the optimized target program should be the same as the original program without optimization.
Optimization usually has several levels (e.g., from \texttt{O0} to \texttt{O3} for the compiler gcc\footnote{https://gcc.gnu.org/}). The higher the optimization level, the greater the difference between the compiled binary code and the source code. Optimization makes it more difficult to generate decompiled code using deep learning models. For example, the optimization could look for redundant operations among lines of code and combine them, or calculate some operations in the compiling time rather than the running time. This would change the structure of HPL. The optimization process is usually irreversible. Also, the optimization strategies are very diverse, even for the same type of operation. For example, in Figure~\ref{o03example}, the target program languages after optimization for the division operations can have different forms when they have different types of operands. Specifically, if the operand is variable, the translated code is \texttt{sdiv}. Moreover, when the operand is changed to immediate, the target code is \texttt{mov, movk, smull, lsr, asr, add}, which does not even contain the division operation.

\vspace{-6pt}
\subsection{Decompilation}
Decompilation is a technique that transforms a compiled executable program or ASM (LPL) into a functionally equivalent HPL~\cite{van2007static}. 
As mentioned previously, to decompile code, analysts would make many heuristic rules to help lift binary code to source code~\cite{Hex-Rays, brumley2013native, kvroustek2017retdec, Ghidra}. However, generalizing the rules is challenging since complex instruction set architectures (ISA), code structure\revise{,} and optimization strategies.
Experts need to summarize the code changes brought by many optimization strategies and handcraft the corresponding decompilation rules. 
For example, in Figure~\ref{o03example}, using different optimization levels, the operation \texttt{var1=var1/-123;} could be compiled into different types of instructions, e.g., \texttt{mov, movk, smull, lsr, asr, add}. 
In this case, the developer needs to handcraft the rules to analyze the data dependencies of these instructions, and determine whether these instructions represent a division statement.
The situation worsens when new operations are added, or new optimization strategies are developed, introducing new rules and impacting the old ones. What is more, this may mislead a neural-based decompiler to translate the similar code \texttt{smull, lsr, asr, add}\revise{,} which does not mean division operation to \texttt{sdiv}. Obtaining a good model requires training on a large-scale dataset with high-quality labels. However, directly using the source code as the label for optimized binary decompilation is inaccurate due to the gap between the source code and the optimized code.

Existing decompilation tools typically design an intermediate representation (IR) as a bridge between the LPL and the HPL. 
While converting IR to HPL is a relatively easy task~\cite{idawhitepaper}, the rules for translating LPL to IR rely on expert analysis and definitions. Moreover, as each tool proposes its own IR and has different definitions of micro-operations, rule-based decompilers suffer from poor generalizability and scalability.
To solve these issues, researchers proposed neural-based decompilation~\cite{katz2018using, katz2019towards, fu2019coda, liang2021neutron}. Katz et al.~\cite{katz2018using} adopted methods from NMT and formulated decompilation as a language translation task, aiming to overcome the bottleneck of rule-based approaches. Coda~\cite{katz2019towards} and Neutron~\cite{liang2021neutron} are designed to learn the mapping rules automatically. The neural-based approaches bring a new idea to program decompilation. However, previous neural-based methods all learn a direct mapping from LPL to HPL or the abstract syntax trees (ASTs) of HPL. 
In addition, none of the current neural-based approaches can handle optimized code, mainly due to the unavailability of a high-quality dataset of optimized code.


\begin{figure*}[!ht]
\centering
\includegraphics[width=0.85\textwidth]{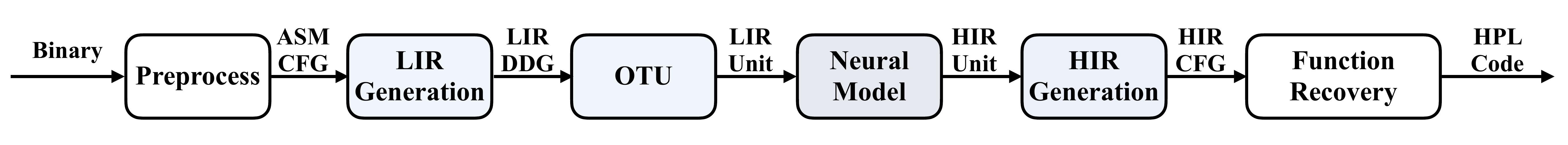}
\vspace{-5pt}
\caption{\textbf{Overview of \toolName}}
\label{pic:framework}
\vspace{-4pt}
\end{figure*}

\vspace{-8pt}


\section{Approach}
\label{sec:approach}

We propose a novel neural decompilation approach \textit{\toolName} that can handle compiler-optimized code. \textit{\toolName} first translates LPL to \textit{\hir} using GNN based IR decompiler model, and then recovers \textit{\hir} code to HPL. Below we elaborate on the design of \textit{\toolName}.


\begin{figure*}[!h]
\centering
\includegraphics[width=0.85\textwidth]{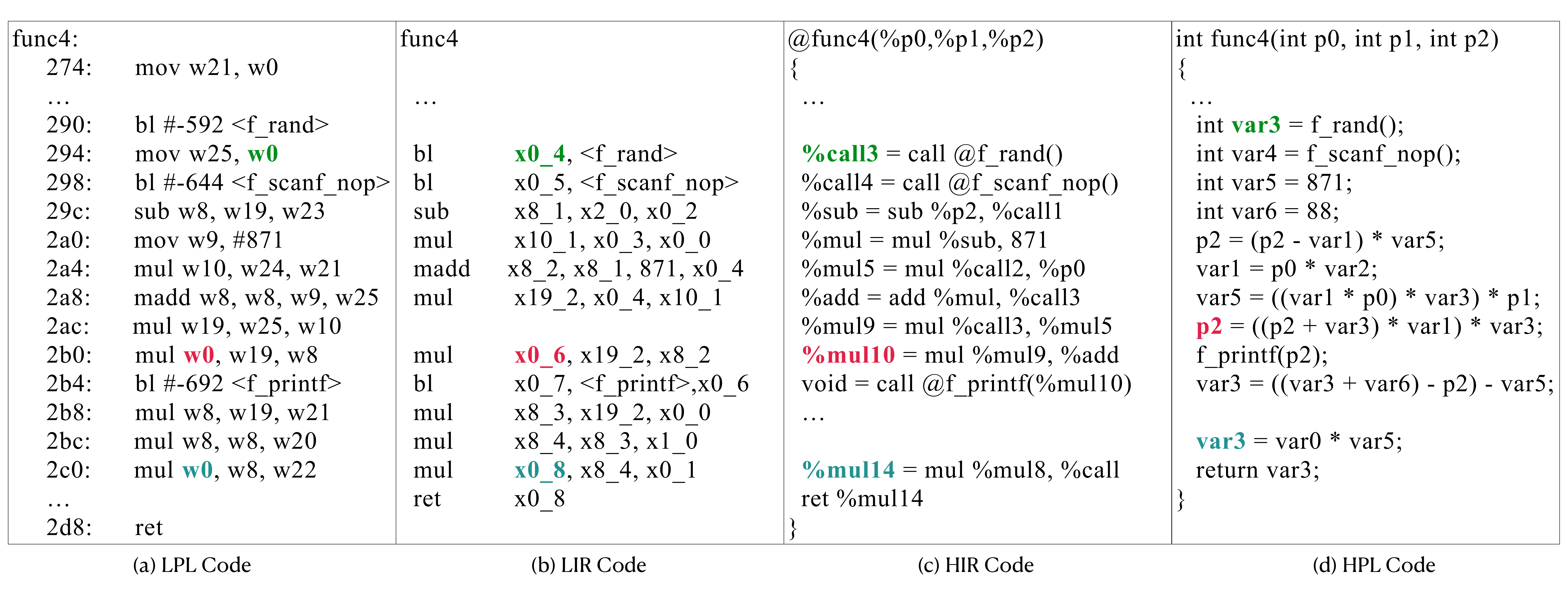}
\vspace{-5pt}
\caption{Example of relationship between HPL, \text{\hir}, \text{\lir}, and LPL} 
\label{whole-example}
\vspace{-8pt}
\end{figure*}

\subsection{Overview} 
\label{subsec-overview}

\begin{figure*}[!h]
\centering
\includegraphics[width=0.85\textwidth]{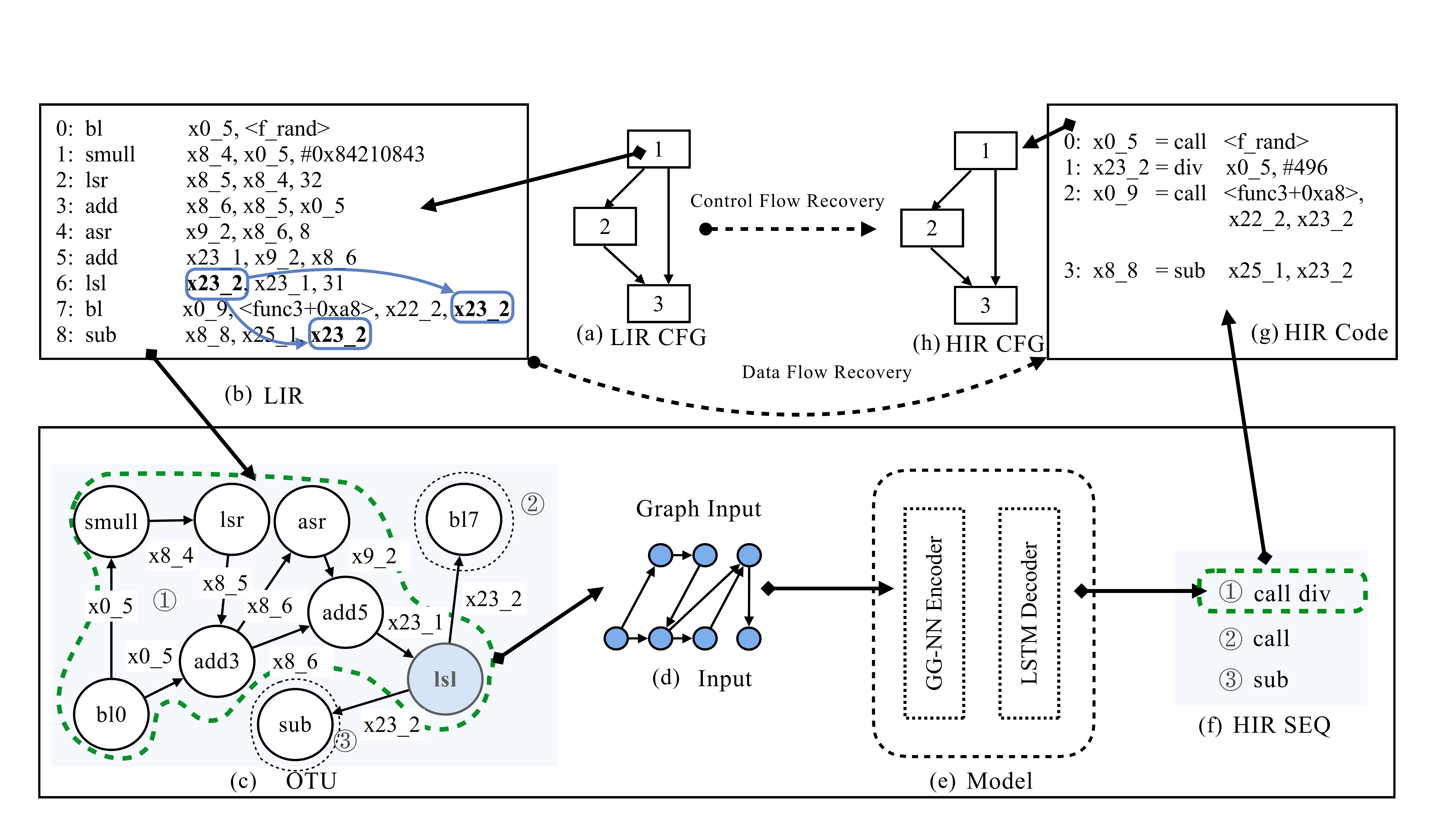}
\vspace{-13pt}
\caption{Neural translation process}
\label{trans-model}
\end{figure*}

The overview of \textit{\toolName} is shown in Figure~\ref{pic:framework}, which aims to decompile the LPL into functionality-equal C-like HPL. And the detail of \textit{\toolName} is shown in Figure~\ref{trans-model}. Considering the large gap between the LPL and HPL, we introduce an IR named \textit{\hir} as a bridge. IR is optimized by the compiler front end. Using the optimized IR as the model's target can reduce the difficulty of model learning since the model no longer needs to learn to reverse the optimization strategies in the compiler's front end. 

In the data construction phase, we first disassemble the binary file, identify the code sections from the binary and retrieve the assembly code of all functions. Then, \textit{\toolName} gets the control flow graph (CFG) for each function and the assembly code of each basic block following previous work~\cite{shoshitaishvili2016state}. 
Next, \textit{\toolName} performs static single assignment (SSA) and data dependency analysis on the LPL in each basic block. We do not use LPL directly as the model's input. Instead, we prefer to use the representation of the SSA form, which is a low-level intermediate representation (\textit{\lir}) (see Section~\ref{subsec-DC}). After generating \textit{\lir}, we further analyze the data dependency between \textit{\lir} code to obtain the data dependency graph (DDG) of \textit{\lir} within the basic block.

\begin{table}
\centering
\footnotesize
\caption{Part of the rules from HIR to HPL}
\vspace{-5pt}
\label{lab:ir2hpl}
\begin{tabular}{m{3.5cm}
<{}|m{3.5cm}
}
\hline
\textbf{HIR}& \textbf{HPL}\\
\hline \hline
\text{\%result = sub \%1, \%2} & \text{result = v1 - v2;}\\ \hline
\text{\%result = add \%1, \%2} & \text{result = v1 + v2;}\\ \hline

\text{\%result = call \@f\_printf, \%1, \%2} & \text{result = f\_printf(v1, v2);}\\ \hline
\text{void = call \@f\_printf, \%1, \%2} & \text{f\_printf(v1, v2);}\\ \hline
\text{ret \%1} & \text{return v1;}\\ \hline
\end{tabular}
\vspace{-8pt}
\end{table}

\begin{table*}[!htbp]
\centering
\footnotesize
\caption{Some instruction templates of \text{\lir}}
\vspace{-5pt}
\label{lab:instruction}
\begin{tabular}{m{4cm}
<{\centering}|m{5cm}
<{}|m{6cm}
<{}}
\hline
\textbf{}& \textbf{LPL}& \textbf{\lir}\\
\hline \hline
\textbf{Return} & \text{\texttt{\textbf{ret}}}& \text{\texttt{\textbf{ret} x0}}\\ \hline

\textbf{Unconditional Branch} & \text{\texttt{\textbf{bl} label}}& \text{\texttt{\textbf{bl} x0, label[, SRC [, SRC...]]}}\\ \hline

\textbf{\makecell*[c]{Conditional Branch}} & \text{\makecell[l]{\texttt{\textbf{cmp} SRC, SRC} \\ \texttt{\textbf{b}.COND label1}}} & 
\text{\texttt{\textbf{b} COND, label1, label2, SRC, SRC}}\\ \hline

\textbf{Store Register} & \text{\texttt{\textbf{str} SRC, DST
}}& \text{\texttt{\textbf{str} DST, SRC
}}\\ \hline

\textbf{Arithmetic (shifted register)} & \text{\texttt{\textbf{add} DST, SRC, SRC, [\textbf{lsl}] IMM}}& \text{\makecell[l]{\texttt{\textbf{add} DST, SRC, SRC} \\ \texttt{\textbf{lsl} DST, IMM}}}\\ \hline

\textbf{Move(wide immediate)} & \text{\makecell[l]{\texttt{\textbf{add} DST, SRC, SRC} \\ \texttt{\textbf{mov} DST, IMM1}}} &\text{\texttt{\textbf{movk} DST, IMM2, [\textbf{lsl}] 16|32) +IMM1}}\\ \hline

\textbf{WZR|XZR Register} & \text{\texttt{\textbf{mov} DST, WZR|XZR IMM}}&\text{\texttt{\textbf{mov} DST, 0}} \\ \hline

\textbf{Conditional Comparison} & \text{\texttt{\textbf{ccmn} SRC, SRC, IMM, cond
}}& \text{\texttt{\textbf{ccmn} nzcv, cond, SRC, SRC, IMM}}\\ \hline
\end{tabular}

\footnotesize{\textbf{\texttt{label}}: jump address. \qquad \textbf{\texttt{label1}}: address of the next instruction. \qquad \textbf{\texttt{IMM}}: immediate. \qquad \textbf{\texttt{nzcv}}: condition flags.}\\
\end{table*}

In the model processing phase, we design a neural model to translate \textit{\lir} into \textit{\hir} (Section~\ref{subcec-Translation}), including the following two steps.
\textbf{\textit{Step 1: Model Training.}} Firstly, \textit{\toolName} prepares a suitable dataset for model training, which contains pairs of \textit{\lir} and \textit{\hir}. \textit{\toolName} splits the basic block into smaller snippets using Optimal Translation Unit (OTU). The \textit{\lir} and \textit{\hir} pair in the corresponding units are functionally equivalent, which makes it easier for the model to learn the transform rules. Secondly, we design and train a neural model based on graph neural network~\cite{2015arXiv151105493L} model to generate \textit{\hir} code.
We describe the detailed design of \textit{\lir}, \textit{\hir} and \textit{\otu} in Section~\ref{subsec-DC} and the construction of training datasets and the model architecture in Section~\ref{subcec-Translation}.
\textbf{\textit{Step 2: Model Translation.}} This step includes the recovery and reorganization of basic blocks in CFG. 
(i) Recovery: \textit{\toolName} recovers statements in basic blocks in this step. Firstly, \textit{\toolName} uses \textit{\otu} to divide the basic block into units and get the DDG of each unit. Then, \textit{\toolName} uses the trained model to translate \textit{\lir} DDG units to \textit{\hir} code templates. After that, \textit{\toolName} fills the analyzed local variables from \textit{\lir} into the \textit{\hir} templates based on data flow analysis. We detail the method of operands recovery in Section~\ref{subsec-postprocessing}. 
(ii) Reorgnization: In this step, we sort these \textit{\hir} snippets based on the position of its corresponding \textit{\lir} in the basic block to recover the complete \textit{\hir} basic block. At last, we recover the CFG of \textit{\hir} (\text{\hir}-CFG) based on the CFG of LPL.

In the HPL generation phase, \textit{\toolName} lifts the \textit{\hir} to HPL. A complete function includes control structures, statements, and function signatures. Translating statements from \textit{\hir} to HPL is not a difficult task, so we make some rules. Table~\ref{lab:ir2hpl} shows some of the rules. For the recovery of control flow and function signatures, many other researchers are focusing on these problems, and we use existing studies~\cite{yakdan2015no,203650} for these two parts. With the function signatures, the statements within each basic block, and the control structures between the basic blocks, we can construct a complete HPL function.

\vspace{-6pt}
\subsection{Dataset Construction}
\label{subsec-DC}

As mentioned previously, we cannot directly use a function’s HPL and LPL as input-output pairs for the model. The main obstacles lie in that (i) the instructions in HPL and LPL have poor correspondence (e.g., redundant or missing operations), \revise{and} (ii) each function has many instructions, making it difficult for the neural network models to learn. In order to solve the problems and facilitate effective learning, we introduce an intermediate representation (\textit{\hir}) that has better instruction correspondence with the LPL and use \textit{\otu} to split the basic blocks into smaller units.

\begin{table*}
\centering
\footnotesize
\caption{HIR Syntax Templates}
\vspace{-5pt}
\label{lab:ir}
\begin{tabular}{m{9cm}
<{}|m{6.5cm}
}
\hline
\textbf{LLVM IR}& \textbf{\text{\toolName} HIR}\\
\hline \hline

\text{<result> = mul <ty> <op1>, <op2>} & \text{<result> = mul <op1>, <op2>}\\ \hline
\text{<result> = add nuw nsw <ty> <op1>, <op2>} & \text{<result> = add <op1>, <op2>}\\ \hline
\text{<result> = fsub [fast-math flags]* <ty> <op1>, <op2>} & \text{<result> = fsub <op1>, <op2>}\\ \hline
\text{<result> = icmp <cond> <ty> <op1>, <op2>} & \text{<result> = icmp <cond> <op1>, <op2>}\\ \hline
\text{switch <intty> <value>, label <defaultdest> [ <intty> <val>, label <dest> ... ]} & \text{switch <value>, <defaultdest> [<val>, <dest> ... ]}\\ \hline

\end{tabular}
\end{table*}

\vspace{1pt}\noindent\textbf{Intermediate Representation.} 
\revise{\textit{\lir} is lifted from LPL by removing machine-related features from LPL, such as registers, designed as the model's input. To get \textit{\lir}, we first change LPL to SSA form, then use optimization strategies similar to constant (register) propagation to eliminate as many registers as possible. \textit{\lir} maintains almost the same syntax as LPL ($<opcode>~<op_{des}>, <op_{src1}>, ...$). Table~\ref{lab:instruction} shows some of the syntax templates of \textit{\lir}. For \textit{\hir}, we extract the operands and opcodes from LLVM IR and rewrite them automatically according to the syntax ($<op_{des}>=<opcode>~<op_{src1}>, ...$), which is a simplified scheme of LLVM IR. It is feasible to directly use LLVM IR instead of HPL as the model's output. However, a model that converts \textit{\lir} to LLVM IR instead of \textit{\hir} needs to learn too much additional information (like data types), which could complicate the model structure and make training such a model extremely difficult. Therefore, we choose to construct the model that converts \textit{\lir} to \textit{\hir}, which is relatively simple (though training this model is still not straightforward).}
\revise{Table~\ref{lab:ir} lists parts of the instruction templates we use for \textit{\hir} and corresponding LLVM IR. Figure~\ref{whole-example}~(c) shows the \textit{\hir} generated by \textit{\toolName}.} 

\begin{figure}[t]
\centering
\includegraphics[width=0.4\textwidth]{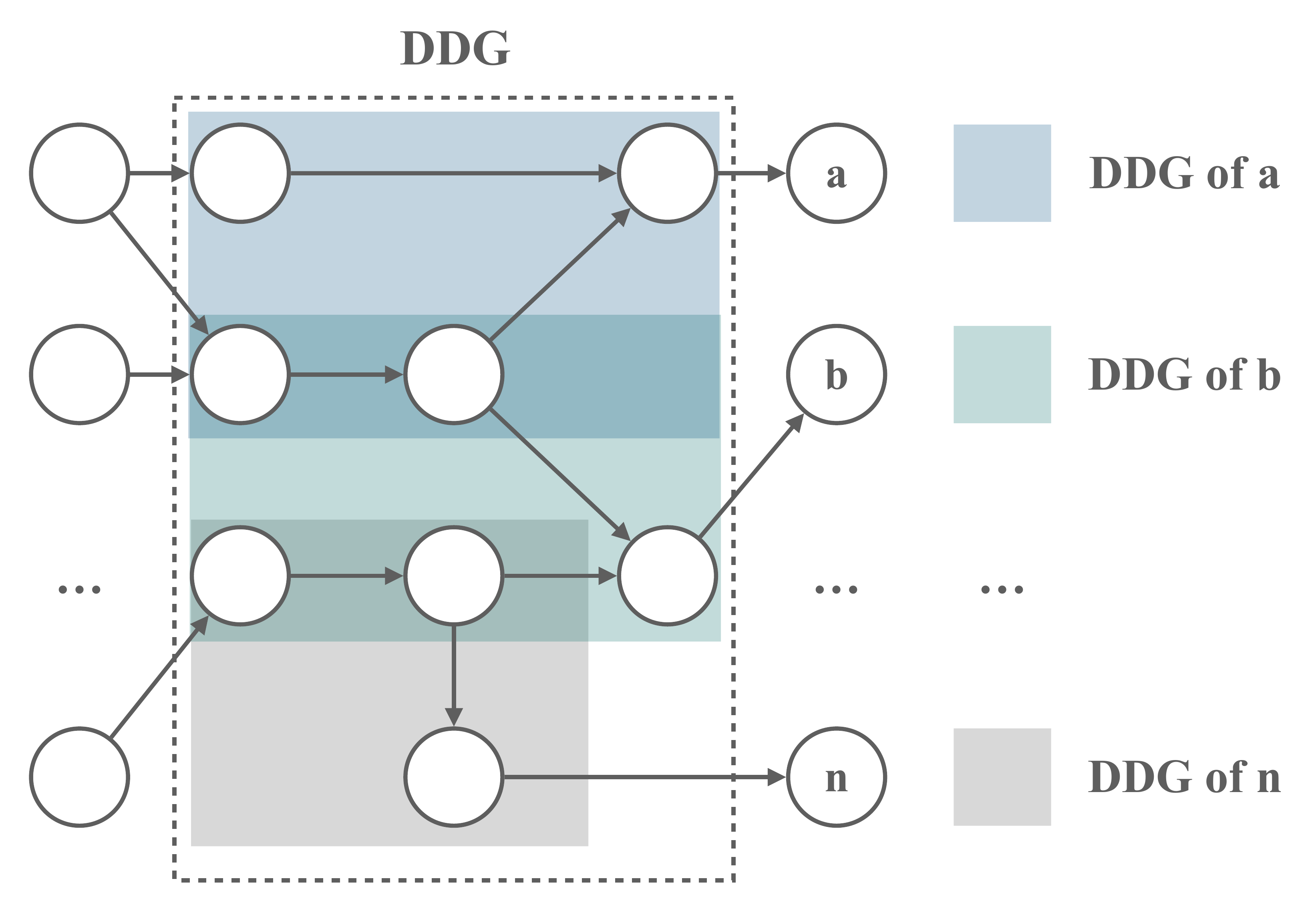}
\vspace{-5pt}
\caption{Black box of basic block} 
\label{bb-blackbox}
\vspace{-10pt}
\end{figure}

\vspace{1pt}\noindent\textbf{Optimal Translation Unit.}
\textit{\toolName} splits the basic block into smaller units that could let the model learn the mapping rules between \textit{\lir} and \textit{\hir} \revise{instructions} easily. An unit in \textit{\lir} should be functionally equivalent to the corresponding unit in \textit{\hir}. One may use a fixed-length translation unit (\tw) to spill a function into units. However, the units generated in this way may not be functionally equivalent. For example, in Figure~\ref{whole-example}, the \texttt{madd} instruction in (b) corresponds to the computation of \texttt{\%mul} and \texttt{\%add} in (c). Nevertheless, the two instructions are located far away and hard to include in a fixed-length \tw. Also, a large size of the \text{\tw} would include unrelated instructions that cannot be paired.

\begin{figure*}[!h]
\centering
\includegraphics[width=0.8\textwidth]{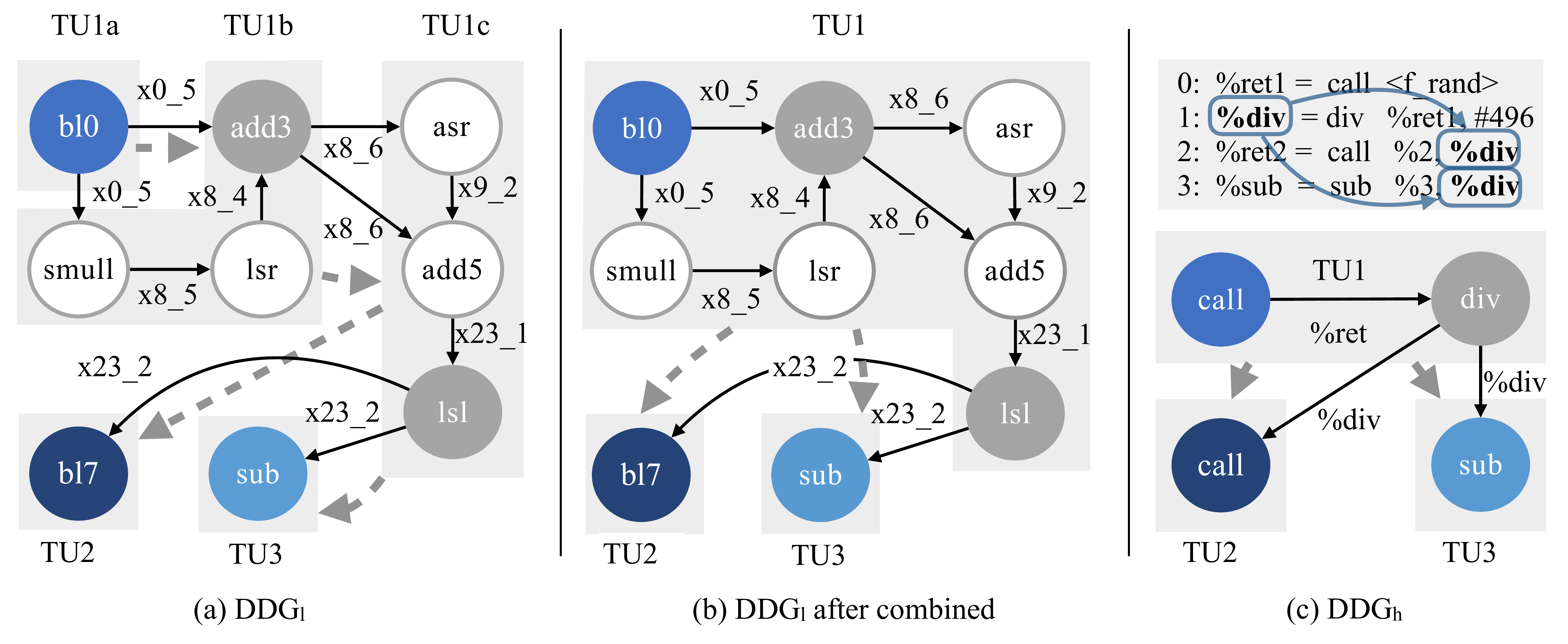}
\vspace{-10pt}
\caption{An example of pair matching of OTU}
\label{fig-DDG-HIR}
\vspace{-8pt}
\end{figure*}

We assume that a basic block is a black box, as shown in Figure~\ref{bb-blackbox}. \del{We do not consider the optimization between basic blocks.}
\del{The reason is that at this stage in the translation process, we only need to translate LIR in every basic block into the corresponding HIR code. This can ensure that the decompilation results are correct without having to worry about combining multiple basic blocks for translation.}
We find that most optimization strategies do not change the output of a basic block. We observe that the output of a basic block usually contains multiple variables whose \revise{data} dependency \del{paths}\revise{graphs} within the basic block often overlap. 
In Figure~\ref{bb-blackbox}, \del{dotted lines}\revise{regions with different colors} corresponding to the variables a, b, and n represent their data dependency \revise{graphs}. To ensure that the optimization to the data dependencies of one variable does not affect the result of other variables, the compiler usually considers optimizing the overlapped \del{parts}and \del{the}independent parts of the \del{DFG}\revise{DDG}, respectively. 
Therefore, we consider that the corresponding parts in \del{DFG}\revise{DDG} of \textit{\hir} and \textit{\lir} has the same semantic. Based on this observation, we design an Optimal Translation Unit (OTU) to divide the overlapping and independent parts of each dependency path of the basic block, which consists of two steps.

Step 1: \textit{\otu} divides a basic block into multiple non-overlapping units. Starting from the input variables of the basic block, \textit{\otu} traverses the entire DDG of the basic block and marks all instructions with two or more out edges as unit boundaries. The \textit{\otu} obtains independent non-overlapping units based on the boundaries. After that, a DDG between units (\text{\cdg}) can be constructed according to the data dependencies between statements. As shown in Figure~\ref{fig-DDG-HIR} (a), the \textit{\otu} first divides the DDG of \textit{\lir} into five non-overlapping units. Each unit is regarded as a node of \text{\cdg}, and the dotted lines are dependency edges. 
In the same way, \del{(c) in}Figure~\ref{fig-DDG-HIR}\revise{~(c)} generates a \text{\cdg} of \textit{\hir}.

Step 2: \textit{\otu} partially merges the units divided in Step 1, because there are units whose out edges all point to one unit. For example, there are 2 out edges between TU1a and TU1b in Figure~\ref{fig-DDG-HIR} (a). This is due to compiler optimizations, such as the division optimization in Figure~\ref{fig-DDG-HIR} that causes operations on a variable to be divided into 3 units TU1a, TU1b, and TU1c.
We iteratively combine such units until no unit in \text{\cdg} has all the out edges pointing to the same unit. For example, in Figure~\ref{fig-DDG-HIR}, (b) is the result of the merging of (a).

\begin{figure}[t]
\centering
\includegraphics[width=0.4\textwidth]{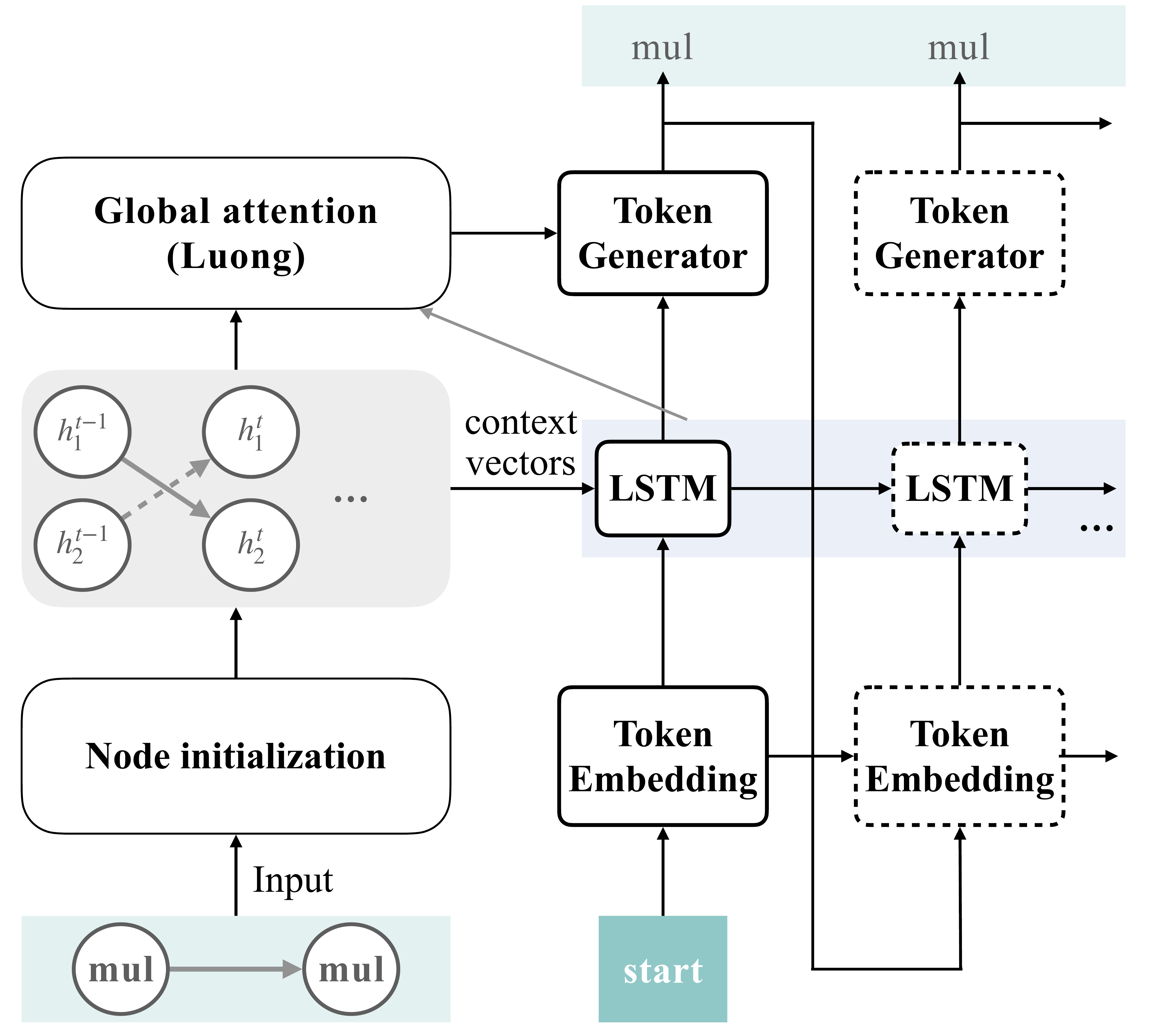}
\caption{Model architecture}
\label{model-arch}
\vspace{-15pt}
\end{figure}

\vspace{2pt}\noindent\textbf{Training Dataset.}
Since the neural model requires labeled data in the training phase, we use \textit{\otu} to partition both the basic blocks of \textit{\lir} and \textit{\hir} when obtaining the training set. Due to the optimization \del{operation}\revise{strategies} of the compiler, the CFG of \textit{\hir} often does not match precisely with the CFG of LPL or \textit{\lir}.
Inaccurate matching between the basic blocks will lead to inaccurate labeling of the training set. 
To avoid this problem, we choose functions that contain only one basic block and then segment them to form the training set of \textit{\toolName} (see Section~\ref{sec:eval}). 
\revise{Based on our observation, optimization across basic blocks only changes the segmentation and their orders, which does not introduce new types of instructions and mappings. Therefore, the model trained on our dataset can accurately translate the \textit{\lir} instructions in each basic block of a complex function to the corresponding \textit{\hir} instructions. 
Existing rule-based decompilers~\cite{Ghidra,Hex-Rays,kvroustek2017retdec} are also implemented based on this principle.}
\del{Since the \textit{\toolName} is translated block-by-block, this dataset does not affect the model's accuracy. }
\del{We choose the DDG of \textit{\lir} as the input to the model and then output the sequence of \textit{\hir} corresponding to the functionality.}

After applying \textit{\otu} in both \textit{\lir} and \textit{\hir}, we try to map units of them to label dataset.
We observe that their \text{\cdg s} are usually isomorphic, although the DDGs of \textit{\lir} and \textit{\hir} may be different. Therefore, we match \text{\cdg s} of \textit{\lir} and \textit{\hir} to pair their units and get labeled. If two nodes in UDG cannot be distinguished, we first examine their internal instructions and distinguish them by some special features, including their constants, const strings, and the address of a procedure call. For nodes that are still indistinguishable according to these features, we throw them away. We use this labeling method to ensure the accuracy of labels.

\vspace{-8pt}
\subsection{Neural Translation} 
\label{subcec-Translation}

In the field of neural translation, sequence-to-sequence (seq2seq) neural networks~\cite{cho2014learning,LSTM,vaswani2017attention} have achieved excellent results and have been applied in commercial products such as Google Translate~\cite{45610}. Therefore, previously studies (e.g., TraFix~\cite{katz2019towards}, Coda~\cite{fu2019coda}, and Neutron~\cite{liang2021neutron}) utilize such models for translation. However, these existing neural machine decompilers do not work well, especially for the optimized LPL, indicating that the seq2seq models cannot effectively cope with the decompilation tasks of LPL. The low accuracy is mainly because seq2seq neural networks do not consider the data dependencies between instructions, which is vital for the compiler to generate LPL. For example, in Figure~\ref{trans-model} (c), the seq2seq neural networks view the \textit{\lir} as the sequence \texttt{<smull, lsr, add3, asr, add5, lsl>}, but ignore the data dependencies (e.g., \texttt{<smull$\rightarrow$lsr>} and \texttt{lsr$\rightarrow$add3>}). So the model wrongly decompiles the result as shifting and arithmetic operations. However, the correct result is the division operation \texttt{div}.   
 
Based on the above observation, the neural network model should capture the instructions and the data dependencies between instructions. So we choose to use Graph Neural Networks (GNN). It can capture the features of nodes (i.e., instructions) and edges (i.e., data dependencies) in DDG. Thus, the model's decompilation problem can be defined as follows: Given the \textit{\lir}'s DDG subgraph $G$ as input, the model outputs the corresponding \textit{\hir} sequence, expressed as $P(Y) = P(Y|G)$.

We adopt the graph-based neural network gated graph sequence neural network (GGS-NN)~\cite{2015arXiv151105493L}. We do not show the details of the model here. Figure~\ref{model-arch} shows the model's architecture (i.e., an encoder-decoder architecture). The encoder uses the gated graph neural network (GG-NN)~\cite{2015arXiv151105493L}. The node initialization module aims to define the initial state of the node and perform initialization operations on the DDG. The decoder uses a long short-term memory (LSTM) network with the bridge mechanism. Considering the inputs are \textit{\lir}/\textit{\hir} units which are not complicated, we use a 2-layer LSTM network. The global attention~\cite{luong2015effective} is introduced to improve the model's performance. The token embedding~\cite{mikolov2013distributed} module is used to generate the word vector of the output sequence, and we use the learnable multi-dimensional embedding vector. Regarding the loss function, we use the Kullback-Leibler divergence~\cite{kullback1951information} as follows.

\begin{align}
\vspace{-20pt}
D_{KL}(p||q) = \sum_{i=1}^n p(x)log\frac{p(x)}{q(x)}
\end{align}

Based on our evaluation, our model is much more accurate (29.58\% higher on average) than seq2seq neural networks. We further look into the code and find that GNN correctly captures the features of data dependencies. For the example in Figure~\ref{trans-model}, our model can correctly decompile the \textit{\lir} to \texttt{div}, which means our model is effective even for optimized code.


\subsection{\textcolor{black}{Operands Recovery}}
\label{subsec-postprocessing}

Recall that the output of our model does not contain real operands.
To accurately recover the \textit{\hir} operands, we further split the \textit{\lir} unit, pair \textit{\lir}/\textit{\hir} instructions, and recover operands in each unit. In this step, we use operands in LIR to fill into the HIR. \del{Particularly, }\revise{We}\del{we} first pair the instructions with the same semantic meanings, which can be obtained by analyzing the instructions manually. For example, in Figure~\ref{trans-model}, the \textit{\lir} instruction \texttt{bl} has the same semantic meaning as the \textit{\hir} instruction \texttt{call}. So we pair them together. Note that identifying the semantic meaning of instructions is only a one-time effort. Then, for the unpaired instructions, we pair them in the order of their addresses. For example, in Figure~\ref{trans-model}, the \textit{\lir} instructions between Line 1 and Line 6 are paired to the \textit{\hir} instruction \texttt{div}.

After obtaining the instruction pairs, we design a data-flow-based approach to recover the \textit{\hir} operands. For each pair, we \del{compare the DDGs. We }identify the destination operand in \textit{\lir} (from the node that has no output link in DDG) and use it as the destination operand in the \textit{\hir} instruction. Then we identify the source operands in \textit{\lir} (from the node that has no parents in DDG) and put them as the source operands in the corresponding \textit{\hir} instruction. \del{For example, for the \textit{\lir} instruction at Line 7 (in Figure~\ref{trans-model}), the generated \textit{\hir} instruction is \texttt{call x0\_9, x22\_2, x23\_2}. Till now, the \textit{\hir} instructions with operands have been generated. }\revise{For some special instructions in \textit{\hir} and \textit{\lir} (e.g., \texttt{div} and \texttt{madd}), we make rules to find the source operands and the destination operands. For example, in Figure~\ref{trans-model}, we map instructions 1-6 (18 operands) in the \textit{\lir} to instruction 1 (3 operands) in the \textit{\hir}. By analyzing the DDG of \textit{\lir}, we get the output variable $x23\_2$, input variable $x0\_5$, and 4 immediate $\#0x84210843$, $32$, $8$, and $31$, which are not defined inside the DDG. Operands $x23\_2$ and $x0\_5$ can be assigned to \textit{\hir} to the corresponding position. Besides, we manually make the division optimization rules to get another operand $\#496$. }

\vspace{-4pt}

\section{Evaluation}
\label{sec:eval}

In this section, we describe our experiments to evaluate \textit{\toolName}'s performance. Firstly, we evaluate the accuracy of decompilation tools at different optimization levels, which can reflect the ability of each decompiler to respond to the optimization strategies related to the expression and data flow in the compiler optimization.
We compare \textit{\toolName} with two state-of-the-art neural-based decompilers~\cite{fu2019coda,liang2021neutron}. To evaluate the efficiency of our model and \textit{\otu}, we compare \textit{\toolName} with other baseline models and other methods of splitting basic blocks.
We also analyze the decompiling results of one famous open-source decompilation tool RetDec~\cite{kvroustek2017retdec}. 


\subsection{Experiment Setup}
\label{subsec-experiment}

\noindent\textbf{Dataset.} 
To build the dataset, we randomly generated 20,000 functions, consisting of arithmetic and calling statements, compiled them using \textit{clang10.0} with optimization levels \texttt{O0} to \texttt{O3}. By capturing the intermediate results and reversing the binaries, we got 80,000 \textit{\lir}/\textit{\hir} pairs. Then, we use \textit{\otu} to split the function into smaller units for training. After removing the duplicated units and batches that were not full, we got 242,000 pairs. We randomly selected 220,000 pairs for training, and the rest 22,000 pairs were used for validation. We evaluate \textit{\toolName} from the following aspects: accuracy and generalizability.

\vspace{1pt}\noindent\textbf{Platform.} All our experiments are conducted on a 64-bit server running Ubuntu 18.04 with 16 cores (Intel(R) Xeon(R) CPU E5-2620 v4 @ 2.10GHz), 128GB memory, 2TB hard drive and 2 GTX Titan-V GPU.

\vspace{-5pt}
\subsection{Accuracy}
\label{subsec:Accuracy}

\noindent\textbf{Metrics.}
As mentioned above, the compiler's optimization changes the statements in the source code. So the decompiled code may not be the same as the original source code at the statements level, even if both codes have the same functionality. We propose a method to evaluate the compiler's accuracy in solving this problem. 
We consider the decompiled basic block correct if the semantics of this HPL's basic block is the same as the semantics of the corresponding LPL's basic block. Below, we will introduce our comparison method for the semantics of two basic blocks.

\del{Basic block represents a series of operations that maps an input set to the output set.} Basic block can be abstracted to a \del{transfer }function $F$, \del{with}\revise{mapping} the input set $IN(B)$ \del{and}\revise{to} \revise{the }output set $OUT(B)$. We define them as follows:
$IN(B) =\{in_0,in_1,...,in_n\}$,
where $in_i$ is the $i$-th input of a basic block.
$OUT(B) =\{out_0, out_1, ...,out_n\}$,
where $out_i$ is the $i$-th output of a basic block.
$F =\{f_0,f_1,...,f_n\}$,
where $f_i$ is the \revise{$i$-th} \del{transfer }function of \revise{a basic block}\del{$out_i$}\revise{,}\del{.}
$out_i = f_i(IN(B))$,
$OUT(B) = F(IN(B))$\revise{.}\del{,}
We define the accuracy of a basic block as \del{$Acc_B = Count_{correct}(f^{HPL}_i)/Count(F_{LPL})$} \revise{$Acc_B = Count_{correct}(f^{HPL}_i)/Count(F^{LPL})$}, where $f^{HPL}_i$ is the $i$-th \del{transfer }function of HPL.
To find the correct $f_i$, we first locate the corresponding \revise{$out_i^{HPL}$}\del{output (e.g., pointer to the same variable)} in the decompiled HPL for each $out_i\revise{^{LPL}}$ in LPL \revise{(e.g., pointer to the same variable)}\del{ and build a possible correct output set $OUT^{'}(B)$}. 
For \textit{\toolName}, it is easy to determine whether the two outputs correspond or not since we map the variables in \revise{\textit{\lir}}\del{LPL} directly to \revise{\textit{\hir}}\del{HPL} when recovering the variables in Section~\ref{sec:approach}. 
For other \del{work}\revise{decompilers}, we pair outputs\del{judge this} by manual analysis.
\revise{Then we can get corresponding $<out_i^{HPL}, out_i^{LPL}>$ pairs.}
Given $IN(B)$, we consider the $f_i^{HPL}$ obtained by decompiling to be correct if the results of the \del{transfer }function $f_i^{HPL}$ and $f_i^{LPL}$ are equal for each \revise{paired $out_i^{HPL}$ and $out_i^{LPL}$}\del{$out_i^{'}$ in $OUT^{'}(B)$}.
At last, we define the \textit{program}  \textit{accuracy} as $Acc = \sum_{k=1}^N Acc_B(B_i)/N$, where $N$ is the number of basic blocks in the program, $B_i$ is the $i$-th basic block in the program.
\revise{In the evaluation, \textit{\toolName} automatically generates functions from the basic blocks, and we manually check if $f_i^{HPL}$ is correct by comparing the functions from HPL and LPL.}
\revise{For example, in Figure~\ref{o03example}, HPL and LPL codes of \texttt{func1} all contain one basic block. We can get the accuracy of \texttt{func1} $Acc=Acc_B(B)/1$. The output set and input set of $B$ in HPL are $\{ret_{HPL}\}$ and $\{p0\}$. The output set and input set of $B$ in LPL are $\{ret_{LPL}\}$ and $\{w19\}$. Then, we can get output pairs in HPL and LPL $\{<ret_{HPL}, ret_{LPL}>\}$. The corresponding function of $ret_{HPL}$ is $f=(f\_scanf\_nop() + p0\times f\_rand())/f\_scanf\_nop()/(-123)$. And the corresponding function of $ret_{LPL}$ is $f=(f\_scanf\_nop()+w19\times f\_rand())/f\_scanf_\_nop()/(-123)$. Note that here we should make rules to handle the division optimization for LPL. We can manually compare these two functions. At last, we can get the accuracy of \texttt{func1} $Acc=Acc_B(B)/1=(1/1)/1=1$, where $Count_{correct}(f^{HPL})=1$ and $Count(F^{LPL})=1$.}


\vspace{1pt}\noindent\textbf{Settings.}
To evaluate the accuracy of \textit{\toolName}, We develop a tool using cfile~\cite{cfile}
to randomly generate 1,000 pieces of code as \textit{DS1}, and then use Clang to compile them at optimization level \texttt{O0-O3} to get 4,000 executable and linkable format files (ELF), where each ELF contains 5 functions. Our dataset includes arithmetic expressions, procedure calls, etc.
We evaluate the robustness of the decompilation tools through their performance in decompiling binary files with (or without) symbolic information, and different optimization levels. Further, we perform strip~\cite{strip} operations on the binary code in the above dataset, where \textit{strip debug} refers to removing the debugging information from the binary, \textit{strip all} means removing all symbolic information from the binary code.

\noindent\textbf{Accuracy of \toolName.}
Table~\ref{tab:performance-opt} shows the accuracy of \textit{\toolName} for four optimization levels from \texttt{O0-O3}. \textit{\toolName} can achieve 92.42\% accuracy on average at \texttt{O0-O3} optimization level. It can be seen that the accuracy of \textit{\toolName} is higher under levels \texttt{O0} and \texttt{O1}, while it is lower at levels \texttt{O2} and \texttt{O3}. At \texttt{O2} and \texttt{O3}, there are more optimizations of the compiler's backend, and the model is more difficult to learn the rules. When the optimization level is increased, the accuracy of the disassembly will reduce, which affects the accuracy of decompilation. 
Under the same optimization level, there are a few differences in the model's performance with and without symbolic information, which indicates that the model has good robustness to the stripped binaries. 
Under level \texttt{O1}, the accuracy rates of binaries without debug information and any symbol tables are slightly lower than with symbolic information. After the analysis, we found that the disassembly accuracy without symbolic information reduces, and some function boundary recognition errors occurred, leading to unsatisfactory decompilation results.

\begin{table}
\centering
\footnotesize
\caption{Results on different compiler-optimization level}
\vspace{-5pt}
\label{tab:performance-opt}
\begin{center}
\begin{tabular}{m{1cm}
<{\centering}|m{2cm}
<{\centering}|m{1cm}
<{\centering}|m{1cm}
<{\centering}|m{1cm}
<{\centering}}
\hline
\multirow{2}*{\textbf{\makecell[c]{Strip\\ option}}} &  \multicolumn{4}{c}{\textbf{Compiler optimization level}} \\
\cline{2-5}
~& \textbf{\texttt{O0}} & \textbf{\texttt{O1}} & \textbf{\texttt{O2}} & \textbf{\texttt{O3}}~\\
\hline
\hline

{\textbf{no}} &0.95 & 1 & 0.9 & 0.9\\
\hline
{\textbf{debug}} &0.95& 0.92 & 0.9 & 0.9\\
\hline
{\textbf{all}} &0.95 & 0.92 & 0.9 & 0.9\\
\hline
\end{tabular}
\end{center}
\footnotesize{\textbf{\texttt{no}}: remain all symbolic information.}\\
\footnotesize{\textbf{\texttt{debug}}: strip debug information.}\\
\footnotesize{\textbf{\texttt{all}}: strip all symbolic information.}\\
\vspace{-15pt}
\end{table}

\begin{table}
\centering
\footnotesize
\caption{Comparison with state-of-the-arts}
\vspace{-5pt}
\label{tab:diff-tool}
\begin{center}
\begin{tabular}{m{1cm}
<{\centering}|m{2cm}
<{\centering}|m{1cm}
<{\centering}|m{1cm}
<{\centering}|m{1cm}
<{\centering}}
\hline
\multirow{2}*{\textbf{}} &  \multicolumn{4}{c}{\textbf{Compiler optimization level}} \\
\cline{2-5}
~& \textbf{\texttt{O0}} & \textbf{\texttt{O1}} & \textbf{\texttt{O2}} & \textbf{\texttt{O3}}~\\
\hline
\hline

{\textbf{Neutron}} &87.78\% & 35.45\% & 32.79\% & 32.81\%\\
\hline
{\textbf{Coda}} &67.2\%-89.2\%* & - & - & -\\
\hline
{\textbf{RetDec}} &29\% & 25\% & 4\% & -\\
\hline
{\textbf{\toolName}} &\textbf{95}\% & \textbf{94.67}\% & \textbf{90}\% & \textbf{90}\%\\
\hline
\end{tabular}
\end{center}
\footnotesize{*Program accuracy 67.2\%-89.2\% of Coda is from Table 2 in ~\cite{fu2019coda} .}\\
\vspace{-10pt}
\end{table}

\vspace{1pt}\noindent\textbf{Comparison with State-of-the-arts.}
We compare \textit{\toolName} with the state-of-the-art neural-based decompilers (i.e., Coda~\cite{fu2019coda} and Neutron~\cite{liang2021neutron}). From the results, we see that \textit{\toolName} outperforms both of them.
We do not have access to the source code and dataset of Coda~\cite{fu2019coda}. We also find that the details needed to reproduce Coda are not described in their paper. So we could neither test Coda on our dataset nor test our \textit{\toolName} on their dataset. 
Considering that the benchmark $(Math+NE)$~\cite{fu2019coda} in Coda is generated \del{in a similar way}\revise{similarly} to our dataset, we directly compare the effect with that described in their paper. Coda splits the long and short sentences in the dataset into two groups for testing $(Math+NE)_S$ and $(Math+NE)_L$. Therefore, the range of Coda's accuracy on these two datasets is listed in Table~\ref{tab:diff-tool}. In addition, since Coda cannot handle the compiler-optimized LPL, this part of the data is replaced with blanks.
For Neutron, we get the code and test it on our dataset.  The result in Table~\ref{tab:diff-tool} shows that Neutron does not perform as well as \textit{\toolName} on our dataset, especially \del{the}\revise{for} compiler-optimized code.
We further analyze the experimental results, where \textit{\toolName} adopts \textit{\hir} as the model's target to cope with compiler optimization and uses the \textit{\otu} mechanism to divide the basic blocks into finer-grained partitions. In contrast, the previous neural-based work utilizes HPL or AST as the model's target. 
Source code and AST are not optimized by the compiler front-end and cannot correspond well with the optimized LPL, increasing the difficulty of model learning.
Therefore, Coda and Neutron cannot cope with the optimized code very well.

\begin{table}
\centering
\footnotesize
\caption{Token accuracy of different neural networks}
\vspace{-5pt}
\label{tab:diff-model}
\begin{center}
\begin{tabular}{m{2.6cm}
<{\centering}|m{0.9cm}
<{\centering}|m{0.9cm}
<{\centering}|m{0.9cm}
<{\centering}|m{0.9cm}
<{\centering}}
\hline
\multirow{2}*{\textbf{}} &  \multicolumn{4}{c}{\textbf{Compiler optimization level}} \\
\cline{2-5}
~& \textbf{\texttt{O0}} & \textbf{\texttt{O1}} & \textbf{\texttt{O2}} & \textbf{\texttt{O3}}~\\
\hline
\hline
{\textbf{Transformer-\texttt{SRC}}} &52.93\% & 38.22\% & 39.84\% & 33.37\%\\
\hline
{\textbf{Transformer-\texttt{AST}}} &75.79\% & 45.62\% & 44.60\% & 32.98\%\\
\hline
{\textbf{Transformer-\texttt{IR}}} &77.18\% & 75.66\% & 74.49\% & 73.94\%\\
\hline
{\textbf{LSTM-\texttt{IR}}} &85.40\% & 86.61\% & 85.63\% & 85.95\%\\
\hline
{\textbf{GRU-\texttt{IR}}} &26.56\% & 21.93\% & 25.85\% & 24.19\%\\
\hline
{\textbf{\toolName}} &\textbf{89.50}\% & \textbf{93.16}\% & \textbf{93.40}\% & \textbf{90.08}\%\\
\hline
\end{tabular}
\end{center}
\vspace{-10pt}
\end{table}

\vspace{1pt}\noindent\textbf{Compare with Different Neural Networks.}
To understand the effect of the GGS-NN model, we compare \textit{\toolName} with other models. We select three wildly used seq2seq models (Transformer~\cite{vaswani2017attention}, LSTM~\cite{LSTM}, and GRU~\cite{cho2014learning}) to make a comparison with our \textit{\toolName}. Note that, instead of using a tree decoder, we serialize (traverse) the AST of the source code as the output of the transformer for model Transformer (AST) in Table~\ref{tab:diff-model}. We use the same data set as \textit{\toolName} to train these models separately. Note that we use Clang to extract \texttt{<assembly, source code/AST/IR>} as ground truth for these models. In this experiment, we use the accuracy of tokens to evaluate these models. This is because outputs of other models often have so many syntax errors that it is hard to evaluate their functionality. The experimental result proves that the GGS-NN can make decompilation results more accurate (see Section~\ref{subcec-Translation}).

We further evaluate the effectiveness of the NMT model using \textit{\hir} as a translation target. We choose the Transformer model as the baseline and use source code, AST, and \textit{\hir} as the model's translation targets for evaluation on DS1. From the first three lines of Table~\ref{tab:diff-model}, we can find that when the NMT model uses source code (SRC) or AST as the translation target (output), its effect on \texttt{O1-O3} is far worse than that at \texttt{O0}. In contrast, the model that uses \textit{\hir} as the output has no significant difference in translation effects at the \texttt{O0-O3}, and the complete accuracy is better than the other two models. The experimental results show that using \textit{\hir} as the translation target of the model is highly generalizable for optimized code, which are not affected by compiler optimizations. 
What's more, using our \textit{\hir} and \textit{\lir} pairs splitting by \textit{\otu} performs better than other models.

\begin{figure}[t]
\centering
\includegraphics[width=0.45\textwidth]{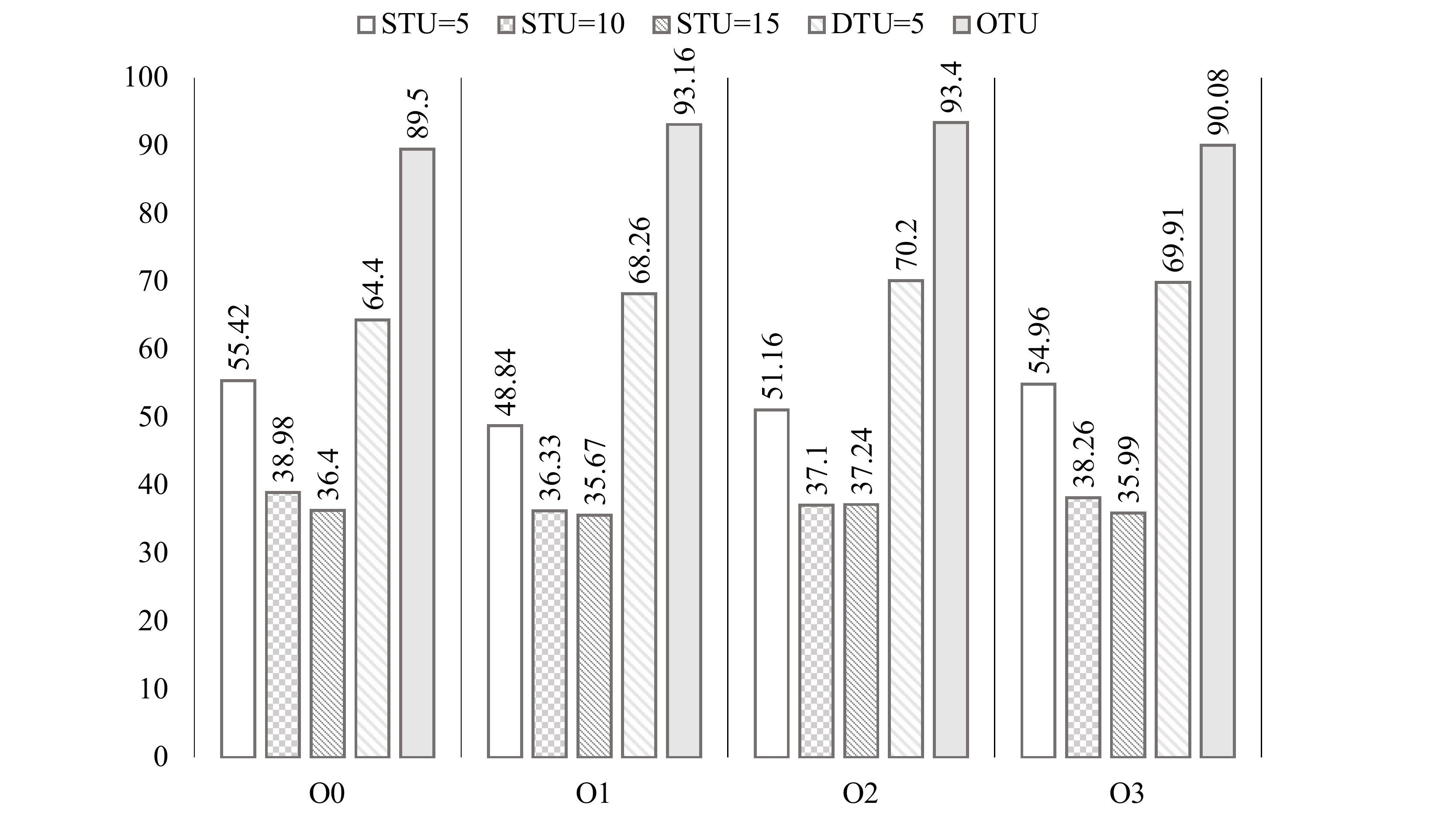}
\vspace{-10pt}
\caption{Token accuracy under different forms of TU}
\label{fig-win}
\vspace{-15pt}
\end{figure}

\vspace{1pt}\noindent\textbf{Impact under Different Translation Unit.} 
We evaluate the token accuracy of using different methods of splitting basic blocks, including code sequence-oriented \textit{\tw} (STU) and DDG-oriented \textit{\tw} (DTU). We select \textit{DS1} to evaluate the performance of different forms of \text{\tw}. To verify the effect of \textit{\tw}, we choose the statement size of the \textit{STU} as 5, 10, and 15 on the assembly sequence. To further verify the performance between fixed-length and variable-length \textit{DTU}, we select a fixed-length \textit{DTU} with a size of 5 and our \textit{\otu}. Figure~\ref{fig-win} shows the performance of \textit{\toolName} under different forms of \textit{TU}. The result indicates that \textit{\toolName} becomes less effective in code sequences as the \textit{STU} increases, mainly because the longer the code the model needs to handle, the more difficult it is to translate accurately. If we do not split the basic block, the results will worsen. 
In addition, we find that the fixed length of \textit{TU}, either sequence-oriented or DDG-oriented, makes the correspondence between \textit{\lir} and \textit{\hir} in the training set more ambiguous, which leads to the model's failure to learn the mapping rules. Compared with the above methods, our \textit{\otu} can maximize the automation of obtaining \textit{\lir} and \textit{\hir} pairs with the correct correspondence for training, thus enabling the model to learn the mapping relationship between them quickly and accurately. What is more worth mentioning is that our approach can deal with compiler optimization problems well.


\vspace{1pt}\noindent\textbf{Analysis of Rule-based Decompilers.}
We also evaluate the performance of one famous open-source rule-based decompiler RetDec~\cite{kvroustek2017retdec}, which contains over 100,000 lines of code and is a representative decompilation work. 
In the evaluation, RetDec does not perform as well as those three neural-based decompilers (only achieving 29\%) on unoptimized binaries. When evaluated on \texttt{O1}, RetDec achieves 29\% accuracy without debug information and 17\% accuracy without any symbolic information. When evaluated on \texttt{O2} and \texttt{O3}, RetDec achieves only 4\% accuracy because mostly binaries are failed to decompile. We manually analyzed these samples of decompilation failures and found that many errors occurred in the disassembly step due to the lack of symbolic information. The code sections could not be accurately located. In some cases, although the function entry point was found, the decompilation is broken due to some small disassembly errors. 
Moreover, we studied the mechanism of rule-based decompilers~\cite{idawhitepaper,Ghidra,kvroustek2017retdec}. Rule-based decompilation tools have more rules and constraints and are more sensitive to disassembly errors. A memory or stack error in the assembly will often cause the following code or the entire function to fail to decompile. In contrast, \textit{\toolName} has no constraints (e.g., stack balance), so it has a certain degree of fault tolerance for some disassembly errors, such as \textit{sp stack unbalanced} caused by inline assembly code. It will not cause the entire decompilation process to fail. Even if the disassembly error leads to the wrong decompilation result, \textit{\toolName} can still finish decompiling without triggering an error and stopping like other tools.

\vspace{-4pt}
\section{Discussion}
\label{sec:discussion}

\noindent\textbf{Limitations.} 
In this work, we propose and implement a novel neural decompilation approach, named \textit{\toolName}, to demonstrate that the neural-based approach copes with the decompilation problem of compiler-optimized LPL. However, \textit{\toolName} still has some limitations. 
Firstly, the HPL statements generated by \textit{\toolName} are mapped directly from \textit{\hir} and are mostly monadic or binary statements. For example, for an expression \texttt{a = b + c * d}, \textit{\toolName}'s outputs are two statements \texttt{tmp = c * d, a = b + tmp}. Such problems could be solved through data dependency analysis.
Besides, the quality of HPL decompiled by \textit{\toolName} is closely related to the accuracy of the LPL (assemble code). We find that for stripped binaries, the disassembly tools (e.g., RetDec) may cause errors in assembly code due to incorrect function boundary identification, especially for decompiled code, which affects the quality of our \textit{\toolName}'s performance.
Secondly, \textit{\toolName} is not completely end-to-end from LPL to HPL. The lifting from IR to HPL depends on the rules, so it is not easy to support multi-machine and multi-language. 
Thirdly, \textit{\toolName} is a prototype system, and the dataset contains only the statements consisting of arithmetic and calling operations to integer variables. We plan to include more types of statements in future work.
Finally, \textit{\toolName} directly uses the existing techniques, including disassembly, reconstruction of control structures, etc. \textit{\toolName} does not consider the errors introduced by these modules so these components will affect the final accuracy.

\vspace{1pt}\noindent\textbf{Future Work.} 
We will continue to explore techniques for improving the decompilation quality of \textit{\toolName} and resolve the above limitations. For example, we will eliminate some \del{of the}binary statements by merging expressions and reducing the redundant variables for HPL generated by \textit{\toolName}. The elimination of sentences will be achieved through data-flow analysis. At the same time, we will use neural-based methods to learn some patterns of statements that match developers' habits through historical experience and guide the process of merging to generate HPL that is more in line with programming habits. Furthermore, we will combine the collective capability of the-state-of-art commercial and open-source disassembly tools~\cite{darki2021disco,flores2020datalog}, to generate high-quality assembly code. The goal is to ensure the \textit{\toolName}'s input is correct, which is a sufficient condition to ensure the performance of the decompiled code.

\vspace{-4pt}
\section{Related Work}
\label{s9}

\noindent\textbf{Rule-based Decompilation.} 
Rule-based decompilation techniques rely on PL experts to customize and design specific heuristic rules lifting low-level PL to high-level PL and achieving software decompilation. The current popular rule-based decompilation tools are Hex-Rays~\cite{Hex-Rays}, RetDec~\cite{kvroustek2017retdec} and Ghidra~\cite{Ghidra}. Hex-Rays is a decompiler engine integrated into the commercial reverse tool IDA Pro, which is the de-facto industry standard in the software security industry. However, Hex-Rays is not open-source, and the inside technology is hard to understand. RetDec is an LLVM-based redirectable open-source decompiler developed by Avast in 2017, aiming to be the first ``universal'' decompiler that can support multiple architectures and languages. RetDec can be used alone or as a plug-in to assist IDA Pro. Ghidra is an SRE framework developed by the National Security Agency (NSA) for cybersecurity missions. 
Ghidra supports running on Windows, macOS, and Linux, supporting multiple processor instruction sets and executable formats. 
Although these studies have made significant improvements, they are far from perfect. Rule-based approaches are needed to manually detect known control flow structures based on written rules and patterns. These rules are difficult to develop, error-prone, usually only capture part of the known CFG, and require long development cycles. Worse still, these methods do not work well when decompiling optimized code. However, optimization is now the default option when commercial software is compiled. Unlike rule-based decompilers, the goal of \textit{\toolName} is based on deep neural networks to learn and extract rules from code data \revise{automatically}. Trying to break through the problem of compiler-optimized code is challenging to decompile accurately.

\vspace{1pt}\noindent\textbf{Learning-based Decompilation.}
Most of the existing learn-based decompilation methods~\cite{katz2018using, katz2019towards, fu2019coda, liang2021neutron} draw on the idea of NMT to transform the decompilation into the problem of mutual translation of two different PL. Katz et al.~\cite{katz2018using} first proposed \revise{an} RNN–based method for decompiling binary code snippets, demonstrating the feasibility of using NMT for decompilation tasks.
Katz et al.~\cite{katz2019towards} proposed a decompilation architecture based on LSTM called TraFix, and they realized that the primary task of building a decompilation tool based on NMT is to make up for the information asymmetry between high-level PL and low-level PL. TraFix takes the preprocessed assembly language as input and the subsequent traversed form of C as output, which reduces the structural asymmetry between the two PLs. Fu et al.~\cite{fu2019coda} proposed an end-to-end neural decompilation framework, named Coda, based on several neural networks and different models used for different statement types. Coda~\cite{fu2019coda} can accurately decompile some simple operations, such as binary operations, which is far from actual application. Unlike the above existing studies, our neural decompilation framework \textit{\toolName} can decompile the real-world low-level PL code, especially the compiler-optimized PL code, into a C-like high-level PL code with corresponding functionality.

\vspace{-4pt}
\section{Conclusions}
\label{sec:conclusion}

In this paper, we propose and implement a neural decompilation framework named \textit{\toolName}, which accurately decompiles LPL code to a C-like HPL with similar functionality. We also design an optimal translation unit (\otu) suitable to form a dataset for learning algorithms better capturing the relationship between HPL and LPL. The evaluation results show that \textit{\toolName} achieves better accuracy for optimized code, even compared with the state-of-the-art.

\vspace{-4pt}
\begin{acks}
This work was supported by NSFC U1836211, Beijing Natural Science Foundation (No.M22004), Youth Innovation Promotion Association CAS, Beijing Academy of Artificial Intelligence (BAAI).
\end{acks}




\begin{thebibliography}{100}
\providecommand{\url}[1]{#1}
\csname url@samestyle\endcsname
\providecommand{\newblock}{\relax}
\providecommand{\bibinfo}[2]{#2}
\providecommand{\BIBentrySTDinterwordspacing}{\spaceskip=0pt\relax}
\providecommand{\BIBentryALTinterwordstretchfactor}{4}
\providecommand{\BIBentryALTinterwordspacing}{\spaceskip=\fontdimen2\font plus
\BIBentryALTinterwordstretchfactor\fontdimen3\font minus
  \fontdimen4\font\relax}
\providecommand{\BIBforeignlanguage}[2]{{%
\expandafter\ifx\csname l@#1\endcsname\relax
\typeout{** WARNING: IEEEtran.bst: No hyphenation pattern has been}%
\typeout{** loaded for the language `#1'. Using the pattern for}%
\typeout{** the default language instead.}%
\else
\language=\csname l@#1\endcsname
\fi
#2}}
\providecommand{\BIBdecl}{\relax}
\BIBdecl

\bibitem{copilot}
``Github copilot,'' 2021, https://copilot.github.com/.

\bibitem{downing_deepreflect_2021}
E.~Downing, Y.~Mirsky, K.~Park, and W.~Lee, ``$\{$DeepReflect$\}$: Discovering
  malicious functionality through binary reconstruction,'' in \emph{30th USENIX
  Security Symposium (USENIX Security 21)}, 2021, pp. 3469--3486.

\bibitem{idawhitepaper}
``Decompiler and beyond,'' 2022, https://infocon.org/cons/.

\bibitem{feng2020codebert}
Z.~Feng, D.~Guo, D.~Tang, N.~Duan, X.~Feng, M.~Gong, L.~Shou, B.~Qin, T.~Liu,
  D.~Jiang \emph{et~al.}, ``Codebert: A pre-trained model for programming and
  natural languages,'' \emph{arXiv preprint arXiv:2002.08155}, 2020.

\bibitem{cifuentes1995decompilation}
C.~Cifuentes and K.~J. Gough, ``Decompilation of binary programs,''
  \emph{Software: Practice and Experience}, vol.~25, no.~7, pp. 811--829, 1995.

\bibitem{liang2021neutron}
R.~Liang, Y.~Cao, P.~Hu, and K.~Chen, ``Neutron: an attention-based neural
  decompiler,'' \emph{Cybersecurity}, vol.~4, no.~1, pp. 1--13, 2021.

\bibitem{cho2014learning}
K.~Cho, B.~Van~Merri{\"e}nboer, C.~Gulcehre, D.~Bahdanau, F.~Bougares,
  H.~Schwenk, and Y.~Bengio, ``Learning phrase representations using rnn
  encoder-decoder for statistical machine translation,'' \emph{arXiv preprint
  arXiv:1406.1078}, 2014.

\bibitem{katz2019towards}
\BIBentryALTinterwordspacing
O.~Katz, Y.~Olshaker, Y.~Goldberg, and E.~Yahav, ``Towards neural
  decompilation,'' \emph{CoRR}, vol. abs/1905.08325, 2019. [Online]. Available:
  \url{http://arxiv.org/abs/1905.08325}
\BIBentrySTDinterwordspacing

\bibitem{fu2019coda}
C.~Fu, H.~Chen, H.~Liu, X.~Chen, Y.~Tian, F.~Koushanfar, and J.~Zhao, ``Coda:
  An end-to-end neural program decompiler,'' in \emph{Advances in Neural
  Information Processing Systems}, 2019, pp. 3703--3714.

\bibitem{Capstone}
``Capstone engine,'' 2019, https://github.com/aquynh/capstone.

\bibitem{shoshitaishvili2016state}
Y.~Shoshitaishvili, R.~Wang, C.~Salls, N.~Stephens, M.~Polino, A.~Dutcher,
  J.~Grosen, S.~Feng, C.~Hauser, C.~Kruegel, and G.~Vigna, ``{SoK: (State of)
  The Art of War: Offensive Techniques in Binary Analysis},'' in \emph{IEEE
  Symposium on Security and Privacy}, 2016.

\bibitem{D-Neliac}
J.~K. Donnelly, ``A decompiler for the countess computer,'' \emph{Navy
  Electronics Laboratory Technical Memorandum 427}, 1960.

\bibitem{10.5555/1096930}
M.~H. Halstead, \emph{Machine-Independent Computer Programming}.\hskip 1em plus
  0.5em minus 0.4em\relax Spartan Books, 1962.

\bibitem{brumley2013native}
D.~Brumley, J.~Lee, E.~J. Schwartz, and M.~Woo, ``Native x86 decompilation
  using semantics-preserving structural analysis and iterative control-flow
  structuring,'' in \emph{Presented as part of the 22nd $\{$USENIX$\}$ Security
  Symposium ($\{$USENIX$\}$ Security 13)}, 2013, pp. 353--368.

\bibitem{Hex-Rays}
``Hex-rays,'' 2021, https://www.hex-rays.com/products/decompiler/.

\bibitem{kvroustek2017retdec}
J.~K{\v{r}}oustek, P.~Matula, and P.~Zemek, ``Retdec: An open-source
  machine-code decompiler,'' 2017.

\bibitem{Ghidra}
``Ghidra,'' 2022, https://ghidra-sre.org/.

\bibitem{lee2011tie}
J.~Lee, T.~Avgerinos, and D.~Brumley, ``Tie: Principled reverse engineering of
  types in binary programs,'' 2011.

\bibitem{lacomis2019dire}
J.~Lacomis, P.~Yin, E.~Schwartz, M.~Allamanis, C.~Le~Goues, G.~Neubig, and
  B.~Vasilescu, ``Dire: A neural approach to decompiled identifier naming,'' in
  \emph{2019 34th IEEE/ACM International Conference on Automated Software
  Engineering (ASE)}.\hskip 1em plus 0.5em minus 0.4em\relax IEEE, 2019, pp.
  628--639.

\bibitem{avast-retargetable}
``Avast retargetable decompiler ida plugin,''
  \url{https://blog.fpmurphy.com/2017/12/avast-retargetable-decompiler-ida-plugin.html},
  2017.

\bibitem{rosenblum2008learning}
N.~E. Rosenblum, X.~Zhu, B.~P. Miller, and K.~Hunt, ``Learning to analyze
  binary computer code.'' in \emph{AAAI}, 2008, pp. 798--804.

\bibitem{karampatziakis2010static}
N.~Karampatziakis, ``Static analysis of binary executables using structural
  svms,'' in \emph{Advances in Neural Information Processing Systems}, 2010,
  pp. 1063--1071.

\bibitem{bao2014byteweight}
T.~Bao, J.~Burket, M.~Woo, R.~Turner, and D.~Brumley, ``$\{$BYTEWEIGHT$\}$:
  Learning to recognize functions in binary code,'' in \emph{23rd
  $\{$USENIX$\}$ Security Symposium ($\{$USENIX$\}$ Security 14)}, 2014, pp.
  845--860.

\bibitem{shin2015recognizing}
E.~C.~R. Shin, D.~Song, and R.~Moazzezi, ``Recognizing functions in binaries
  with neural networks,'' in \emph{24th $\{$USENIX$\}$ Security Symposium
  ($\{$USENIX$\}$ Security 15)}, 2015, pp. 611--626.

\bibitem{203650}
\BIBentryALTinterwordspacing
Z.~L. Chua, S.~Shen, P.~Saxena, and Z.~Liang, ``Neural nets can learn function
  type signatures from binaries,'' in \emph{26th USENIX Security Symposium
  (USENIX Security 17)}.\hskip 1em plus 0.5em minus 0.4em\relax Vancouver, BC:
  USENIX Association, Aug. 2017, pp. 99--116. [Online]. Available:
  \url{https://www.usenix.org/conference/usenixsecurity17/technical-sessions/presentation/chua}
\BIBentrySTDinterwordspacing

\bibitem{chua2017neural}
------, ``Neural nets can learn function type signatures from binaries,'' in
  \emph{26th $\{$USENIX$\}$ Security Symposium ($\{$USENIX$\}$ Security 17)},
  2017, pp. 99--116.

\bibitem{jaffe2018meaningful}
A.~Jaffe, J.~Lacomis, E.~J. Schwartz, C.~L. Goues, and B.~Vasilescu,
  ``Meaningful variable names for decompiled code: A machine translation
  approach,'' in \emph{Proceedings of the 26th Conference on Program
  Comprehension}, 2018, pp. 20--30.

\bibitem{he2018debin}
J.~He, P.~Ivanov, P.~Tsankov, V.~Raychev, and M.~Vechev, ``Debin: Predicting
  debug information in stripped binaries,'' in \emph{Proceedings of the 2018
  ACM SIGSAC Conference on Computer and Communications Security}, 2018, pp.
  1667--1680.

\bibitem{hindle2012naturalness}
A.~Hindle, E.~T. Barr, Z.~Su, M.~Gabel, and P.~Devanbu, ``On the naturalness of
  software,'' in \emph{2012 34th International Conference on Software
  Engineering (ICSE)}.\hskip 1em plus 0.5em minus 0.4em\relax IEEE, 2012, pp.
  837--847.

\bibitem{katz2018using}
D.~S. Katz, J.~Ruchti, and E.~Schulte, ``Using recurrent neural networks for
  decompilation,'' in \emph{2018 IEEE 25th International Conference on Software
  Analysis, Evolution and Reengineering (SANER)}.\hskip 1em plus 0.5em minus
  0.4em\relax IEEE, 2018, pp. 346--356.

\bibitem{kommrusch2020equivalence}
S.~Kommrusch, T.~Barollet, and L.-N. Pouchet, ``Equivalence of dataflow graphs
  via rewrite rules using a graph-to-sequence neural model,'' \emph{arXiv
  preprint arXiv:2002.06799}, 2020.

\bibitem{van2007static}
M.~J. Van~Emmerik, \emph{Static single assignment for decompilation}.\hskip 1em
  plus 0.5em minus 0.4em\relax University of Queensland, 2007.

\bibitem{sutskever2014sequence}
I.~Sutskever, O.~Vinyals, and Q.~V. Le, ``Sequence to sequence learning with
  neural networks,'' in \emph{Advances in neural information processing
  systems}, 2014, pp. 3104--3112.

\bibitem{wu2016google}
Y.~Wu, M.~Schuster, Z.~Chen, Q.~V. Le, M.~Norouzi, W.~Macherey, M.~Krikun,
  Y.~Cao, Q.~Gao, K.~Macherey \emph{et~al.}, ``Google's neural machine
  translation system: Bridging the gap between human and machine translation,''
  \emph{arXiv preprint arXiv:1609.08144}, 2016.

\bibitem{shaw2018self}
P.~Shaw, J.~Uszkoreit, and A.~Vaswani, ``Self-attention with relative position
  representations,'' \emph{arXiv preprint arXiv:1803.02155}, 2018.

\bibitem{bahdanau2014neural}
D.~Bahdanau, K.~Cho, and Y.~Bengio, ``Neural machine translation by jointly
  learning to align and translate,'' \emph{arXiv preprint arXiv:1409.0473},
  2014.

\bibitem{Google-C++-Style-Guide}
``Google c++ style guide,'' 2022,
  https://google.github.io/styleguide/cppguide.html.

\bibitem{Debin}
``Debin,'' 2021, https://debin.ai/.

\bibitem{ding2019asm2vec}
S.~H. Ding, B.~C. Fung, and P.~Charland, ``Asm2vec: Boosting static
  representation robustness for binary clone search against code obfuscation
  and compiler optimization,'' in \emph{2019 IEEE Symposium on Security and
  Privacy (SP)}.\hskip 1em plus 0.5em minus 0.4em\relax IEEE, 2019, pp.
  472--489.

\bibitem{vinyals2015show}
O.~Vinyals, A.~Toshev, S.~Bengio, and D.~Erhan, ``Show and tell: A neural image
  caption generator,'' in \emph{Proceedings of the IEEE conference on computer
  vision and pattern recognition}, 2015, pp. 3156--3164.

\bibitem{schwartz2018using}
E.~J. Schwartz, C.~F. Cohen, M.~Duggan, J.~Gennari, J.~S. Havrilla, and
  C.~Hines, ``Using logic programming to recover c++ classes and methods from
  compiled executables,'' in \emph{Proceedings of the 2018 ACM SIGSAC
  Conference on Computer and Communications Security}, 2018, pp. 426--441.

\bibitem{jin2014recovering}
W.~Jin, C.~Cohen, J.~Gennari, C.~Hines, S.~Chaki, A.~Gurfinkel, J.~Havrilla,
  and P.~Narasimhan, ``Recovering c++ objects from binaries using
  inter-procedural data-flow analysis,'' in \emph{Proceedings of ACM SIGPLAN on
  Program Protection and Reverse Engineering Workshop 2014}, 2014, pp. 1--11.

\bibitem{schkufza2013stochastic}
E.~Schkufza, R.~Sharma, and A.~Aiken, ``Stochastic superoptimization,''
  \emph{ACM SIGARCH Computer Architecture News}, vol.~41, no.~1, pp. 305--316,
  2013.

\bibitem{klein2017opennmt}
G.~Klein, Y.~Kim, Y.~Deng, J.~Senellart, and A.~M. Rush, ``Opennmt: Open-source
  toolkit for neural machine translation,'' \emph{arXiv preprint
  arXiv:1701.02810}, 2017.

\bibitem{warren2013hacker}
H.~S. Warren, \emph{Hacker's delight}.\hskip 1em plus 0.5em minus 0.4em\relax
  Pearson Education, 2013.

\bibitem{Algorithms/C}
``Algorithms/c,'' 2019, https://github.com/Thuva4/Algorithms/tree/ma\\ster/C.

\bibitem{Libpcap}
``Libpcap,'' 2019, https://www.tcpdump.org/.

\bibitem{strip}
``strip,'' 2009, https://linux.die.net/man/1/strip.

\bibitem{GMP}
``Gmp,'' 2019, https://gmplib.org/.

\bibitem{leven}
``Levenshtein,'' \url{https://en.wikipedia.org/wiki/Levenshtein\_distance},
  2021.

\bibitem{liu2020far}
Z.~Liu and S.~Wang, ``How far we have come: Testing decompilation correctness
  of c decompilers,'' in \emph{Proceedings of the 29th ACM SIGSOFT
  International Symposium on Software Testing and Analysis}, 2020, pp.
  475--487.

\bibitem{tian2018attention}
D.~J. Tian, G.~Hernandez, J.~I. Choi, V.~Frost, C.~Raules, P.~Traynor,
  H.~Vijayakumar, L.~Harrison, A.~Rahmati, M.~Grace \emph{et~al.}, ``Attention
  spanned: Comprehensive vulnerability analysis of $\{$AT$\}$ commands within
  the android ecosystem,'' in \emph{27th $\{$USENIX$\}$ Security Symposium
  ($\{$USENIX$\}$ Security 18)}, 2018, pp. 273--290.

\bibitem{hernandez2020bigmac}
G.~Hernandez, D.~J. Tian, A.~S. Yadav, B.~J. Williams, and K.~R. Butler,
  ``Bigmac: Fine-grained policy analysis of android firmware,'' in \emph{29th
  $\{$USENIX$\}$ Security Symposium ($\{$USENIX$\}$ Security 20)}, 2020, pp.
  271--287.

\bibitem{appel2004modern}
A.~W. Appel, \emph{Modern compiler implementation in C}.\hskip 1em plus 0.5em
  minus 0.4em\relax Cambridge university press, 2004.

\bibitem{LSTM}
\BIBentryALTinterwordspacing
S.~Hochreiter and J.~Schmidhuber, ``Long short-term memory,'' \emph{Neural
  Comput.}, vol.~9, no.~8, p. 1735–1780, Nov. 1997. [Online]. Available:
  \url{https://doi.org/10.1162/neco.1997.9.8.1735}
\BIBentrySTDinterwordspacing

\bibitem{45610}
\BIBentryALTinterwordspacing
Y.~Wu, M.~Schuster, Z.~Chen, Q.~V. Le, M.~Norouzi, W.~Macherey, M.~Krikun,
  Y.~Cao, Q.~Gao, K.~Macherey, J.~Klingner, A.~Shah, M.~Johnson, X.~Liu,
  Łukasz Kaiser, S.~Gouws, Y.~Kato, T.~Kudo, H.~Kazawa, K.~Stevens, G.~Kurian,
  N.~Patil, W.~Wang, C.~Young, J.~Smith, J.~Riesa, A.~Rudnick, O.~Vinyals,
  G.~Corrado, M.~Hughes, and J.~Dean, ``Google's neural machine translation
  system: Bridging the gap between human and machine translation,''
  \emph{CoRR}, vol. abs/1609.08144, 2016. [Online]. Available:
  \url{http://arxiv.org/abs/1609.08144}
\BIBentrySTDinterwordspacing

\bibitem{elsabagh2020firmscope}
M.~Elsabagh, R.~Johnson, A.~Stavrou, C.~Zuo, Q.~Zhao, and Z.~Lin,
  ``$\{$FIRMSCOPE$\}$: Automatic uncovering of privilege-escalation
  vulnerabilities in pre-installed apps in android firmware,'' in \emph{29th
  $\{$USENIX$\}$ Security Symposium ($\{$USENIX$\}$ Security 20)}, 2020, pp.
  2379--2396.

\bibitem{yakdan2016helping}
K.~Yakdan, S.~Dechand, E.~Gerhards-Padilla, and M.~Smith, ``Helping johnny to
  analyze malware: A usability-optimized decompiler and malware analysis user
  study,'' in \emph{2016 IEEE Symposium on Security and Privacy (SP)}.\hskip
  1em plus 0.5em minus 0.4em\relax IEEE, 2016, pp. 158--177.

\bibitem{LeetCode-in-pure-C}
``Leetcode in pure c,'' 2021, https://github.com/begeekmyfriend/leetcode.

\bibitem{cfile}
``cfile,'' 2021, https://github.com/cogu/cfile.

\bibitem{Clang}
``Clang,'' 2021, https://clang.llvm.org.

\bibitem{nbref}
``Nbref,'' 2021, https://github.com/facebookresearch/nbref.git.

\bibitem{llvmir}
``llvmir,'' 2021, https://llvm.org/docs/LangRef.html.

\bibitem{luong2015effective}
M.-T. Luong, H.~Pham, and C.~D. Manning, ``Effective approaches to
  attention-based neural machine translation,'' \emph{arXiv preprint
  arXiv:1508.04025}, 2015.

\bibitem{mikolov2013distributed}
T.~Mikolov, I.~Sutskever, K.~Chen, G.~S. Corrado, and J.~Dean, ``Distributed
  representations of words and phrases and their compositionality,'' in
  \emph{Advances in neural information processing systems}, 2013, pp.
  3111--3119.

\bibitem{yakdan2015no}
K.~Yakdan, S.~Eschweiler, E.~Gerhards-Padilla, and M.~Smith, ``No more gotos:
  Decompilation using pattern-independent control-flow structuring and
  semantic-preserving transformations.'' in \emph{NDSS}.\hskip 1em plus 0.5em
  minus 0.4em\relax Citeseer, 2015.

\bibitem{Abril07}
\BIBentryALTinterwordspacing
P.~S. Abril and R.~Plant, ``The patent holder's dilemma: Buy, sell, or troll?''
  \emph{Communications of the ACM}, vol.~50, no.~1, pp. 36--44, Jan. 2007.
  [Online]. Available: \url{http://doi.acm.org/10.1145/1219092.1219093}
\BIBentrySTDinterwordspacing

\bibitem{Cohen07}
\BIBentryALTinterwordspacing
S.~Cohen, W.~Nutt, and Y.~Sagic, ``Deciding equivalances among conjunctive
  aggregate queries,'' \emph{J. ACM}, vol.~54, no.~2, Apr. 2007. [Online].
  Available: \url{http://doi.acm.org/10.1145/1219092.1219093}
\BIBentrySTDinterwordspacing

\bibitem{JCohen96}
J.~Cohen, Ed., \emph{Special issue: Digital Libraries}, vol.~39, no.~11, Nov.
  1996.

\bibitem{Kosiur01}
D.~Kosiur, \emph{Understanding Policy-Based Networking}, 2nd~ed.\hskip 1em plus
  0.5em minus 0.4em\relax New York, NY: Wiley, 2001.

\bibitem{Harel79}
\BIBentryALTinterwordspacing
D.~Harel, \emph{First-Order Dynamic Logic}, ser. Lecture Notes in Computer
  Science.\hskip 1em plus 0.5em minus 0.4em\relax New York, NY:
  Springer-Verlag, 1979, vol.~68. [Online]. Available:
  \url{http://dx.doi.org/10.1007/3-540-09237-4}
\BIBentrySTDinterwordspacing

\bibitem{Editor00}
\BIBentryALTinterwordspacing
I.~Editor, Ed., \emph{The title of book one}, 1st~ed., ser. The name of the
  series one.\hskip 1em plus 0.5em minus 0.4em\relax Chicago: University of
  Chicago Press, 2007, vol.~9. [Online]. Available:
  \url{http://dx.doi.org/10.1007/3-540-09456-9}
\BIBentrySTDinterwordspacing

\bibitem{Editor00a}
\BIBentryALTinterwordspacing
------, \emph{The title of book two}, 2nd~ed., ser. The name of the series
  two.\hskip 1em plus 0.5em minus 0.4em\relax Chicago: University of Chicago
  Press, 2008, ch. 100. [Online]. Available:
  \url{http://dx.doi.org/10.1007/3-540-09456-9}
\BIBentrySTDinterwordspacing

\bibitem{Spector90}
\BIBentryALTinterwordspacing
A.~Z. Spector, ``Achieving application requirements,'' in \emph{Distributed
  Systems}, 2nd~ed., S.~Mullender, Ed.\hskip 1em plus 0.5em minus 0.4em\relax
  New York, NY: ACM Press, 1990, pp. 19--33. [Online]. Available:
  \url{http://doi.acm.org/10.1145/90417.90738}
\BIBentrySTDinterwordspacing

\bibitem{Douglass98}
\BIBentryALTinterwordspacing
B.~P. Douglass, D.~Harel, and M.~B. Trakhtenbrot, ``Statecarts in use:
  structured analysis and object-orientation,'' in \emph{Lectures on Embedded
  Systems}, ser. Lecture Notes in Computer Science, G.~Rozenberg and F.~W.
  Vaandrager, Eds.\hskip 1em plus 0.5em minus 0.4em\relax London:
  Springer-Verlag, 1998, vol. 1494, pp. 368--394. [Online]. Available:
  \url{http://dx.doi.org/10.1007/3-540-65193-4_29}
\BIBentrySTDinterwordspacing

\bibitem{Knuth97}
D.~E. Knuth, \emph{The Art of Computer Programming, Vol. 1: Fundamental
  Algorithms (3rd. ed.)}.\hskip 1em plus 0.5em minus 0.4em\relax Addison Wesley
  Longman Publishing Co., Inc., 1997.

\bibitem{Knuth98}
------, \emph{The Art of Computer Programming}, 3rd~ed., ser. Fundamental
  Algorithms.\hskip 1em plus 0.5em minus 0.4em\relax Addison Wesley Longman
  Publishing Co., Inc., 1998, vol.~1, (book).

\bibitem{GM05}
D.~Geiger and C.~Meek, ``Structured variational inference procedures and their
  realizations (as incol),'' in \emph{Proceedings of Tenth International
  Workshop on Artificial Intelligence and Statistics, {\rm The
  Barbados}}.\hskip 1em plus 0.5em minus 0.4em\relax The Society for Artificial
  Intelligence and Statistics, Jan. 2005.

\bibitem{flores2020datalog}
A.~Flores-Montoya and E.~Schulte, ``Datalog disassembly,'' in \emph{29th USENIX
  Security Symposium (USENIX Security 20)}, 2020, pp. 1075--1092.

\bibitem{Smith10}
\BIBentryALTinterwordspacing
S.~W. Smith, ``An experiment in bibliographic mark-up: Parsing metadata for xml
  export,'' in \emph{Proceedings of the 3rd. annual workshop on Librarians and
  Computers}, ser. LAC '10, R.~N. Smythe and A.~Noble, Eds., vol.~3.\hskip 1em
  plus 0.5em minus 0.4em\relax Milan Italy: Paparazzi Press, 2010, pp.
  422--431. [Online]. Available: \url{http://dx.doi.org/99.0000/woot07-S422}
\BIBentrySTDinterwordspacing

\bibitem{VanGundy07}
M.~V. Gundy, D.~Balzarotti, and G.~Vigna, ``Catch me, if you can: Evading
  network signatures with web-based polymorphic worms,'' in \emph{Proceedings
  of the first USENIX workshop on Offensive Technologies}, ser. WOOT '07.\hskip
  1em plus 0.5em minus 0.4em\relax Berkley, CA: USENIX Association, 2007.

\bibitem{VanGundy08}
------, ``Catch me, if you can: Evading network signatures with web-based
  polymorphic worms,'' in \emph{Proceedings of the first USENIX workshop on
  Offensive Technologies}, ser. WOOT '08.\hskip 1em plus 0.5em minus
  0.4em\relax Berkley, CA: USENIX Association, 2008, pp. 99--100.

\bibitem{VanGundy09}
------, ``Catch me, if you can: Evading network signatures with web-based
  polymorphic worms,'' in \emph{Proceedings of the first USENIX workshop on
  Offensive Technologies}, ser. WOOT '09.\hskip 1em plus 0.5em minus
  0.4em\relax Berkley, CA: USENIX Association, 2009, pp. 90--100.

\bibitem{Andler79}
\BIBentryALTinterwordspacing
S.~Andler, ``Predicate path expressions,'' in \emph{Proceedings of the 6th. ACM
  SIGACT-SIGPLAN symposium on Principles of Programming Languages}, ser. POPL
  '79.\hskip 1em plus 0.5em minus 0.4em\relax New York, NY: ACM Press, 1979,
  pp. 226--236. [Online]. Available:
  \url{http://doi.acm.org/10.1145/567752.567774}
\BIBentrySTDinterwordspacing

\bibitem{Harel78}
D.~Harel, ``Logics of programs: Axiomatics and descriptive power,''
  Massachusetts Institute of Technology, Cambridge, MA, MIT Research Lab
  Technical Report TR-200, 1978.

\bibitem{anisi03}
D.~A. Anisi, ``Optimal motion control of a ground vehicle,'' Master's thesis,
  Royal Institute of Technology (KTH), Stockholm, Sweden, 2003.

\bibitem{Clarkson85}
K.~L. Clarkson, ``Algorithms for closest-point problems (computational
  geometry),'' Ph.D. dissertation, Stanford University, Palo Alto, CA, 1985,
  uMI Order Number: AAT 8506171.

\bibitem{Thornburg01}
\BIBentryALTinterwordspacing
H.~Thornburg. (2001, Mar.) Introduction to bayesian statistics. [Online].
  Available: \url{http://ccrma.stanford.edu/~jos/bayes/bayes.html}
\BIBentrySTDinterwordspacing

\bibitem{Ablamowicz07}
\BIBentryALTinterwordspacing
R.~Ablamowicz and B.~Fauser. (2007) Clifford: a maple 11 package for clifford
  algebra computations, version 11. [Online]. Available:
  \url{http://math.tntech.edu/rafal/cliff11/index.html}
\BIBentrySTDinterwordspacing

\bibitem{Poker06}
\BIBentryALTinterwordspacing
Poker-Edge.Com, ``Stats and analysis,'' Mar. 2006. [Online]. Available:
  \url{http://www.poker-edge.com/stats.php}
\BIBentrySTDinterwordspacing

\bibitem{Obama08}
\BIBentryALTinterwordspacing
B.~Obama, ``A more perfect union,'' Video, Mar. 2008. [Online]. Available:
  \url{http://video.google.com/videoplay?docid=6528042696351994555}
\BIBentrySTDinterwordspacing

\bibitem{JoeScientist001}
J.~Scientist, ``The fountain of youth,'' Aug. 2009, patent No. 12345, Filed
  July 1st., 2008, Issued Aug. 9th., 2009.

\bibitem{Novak03}
\BIBentryALTinterwordspacing
D.~Novak, ``Solder man,'' in \emph{ACM SIGGRAPH 2003 Video Review on Animation
  theater Program: Part I - Vol. 145 (July 27--27, 2003)}.\hskip 1em plus 0.5em
  minus 0.4em\relax New York, NY: ACM Press, March 21, 2008 2003, p.~4.
  [Online]. Available:
  \url{http://video.google.com/videoplay?docid=6528042696351994555}
\BIBentrySTDinterwordspacing

\bibitem{Lee05}
\BIBentryALTinterwordspacing
N.~Lee, ``Interview with bill kinder: January 13, 2005,'' \emph{Comput.
  Entertain.}, vol.~3, no.~1, Jan.-March 2005. [Online]. Available:
  \url{http://doi.acm.org/10.1145/1057270.1057278}
\BIBentrySTDinterwordspacing

\bibitem{Rous08}
B.~Rous, ``The enabling of digital libraries,'' \emph{Digital Libraries},
  vol.~12, no.~3, Jul. 2008, to appear.

\bibitem{384253}
R.~Werneck, J.~a. Setubal, and A.~da~Conceic\~{a}o, ``(old) finding minimum
  congestion spanning trees,'' \emph{J. Exp. Algorithmics}, vol.~5, p.~11,
  2000.

\bibitem{Werneck:2000:FMC:351827.384253}
\BIBentryALTinterwordspacing
------, ``(new) finding minimum congestion spanning trees,'' \emph{J. Exp.
  Algorithmics}, vol.~5, Dec. 2000. [Online]. Available:
  \url{http://portal.acm.org/citation.cfm?id=351827.384253}
\BIBentrySTDinterwordspacing

\bibitem{1555162}
M.~Conti, R.~Di~Pietro, L.~V. Mancini, and A.~Mei, ``(old) distributed data
  source verification in wireless sensor networks,'' \emph{Inf. Fusion},
  vol.~10, no.~4, pp. 342--353, 2009.

\bibitem{Conti:2009:DDS:1555009.1555162}
\BIBentryALTinterwordspacing
------, ``(new) distributed data source verification in wireless sensor
  networks,'' \emph{Inf. Fusion}, vol.~10, no.~4, pp. 342--353, Oct. 2009.
  [Online]. Available:
  \url{http://portal.acm.org/citation.cfm?id=1555009.1555162}
\BIBentrySTDinterwordspacing

\bibitem{Li:2008:PUC:1358628.1358946}
\BIBentryALTinterwordspacing
C.-L. Li, A.~G. Buyuktur, D.~K. Hutchful, N.~B. Sant, and S.~K. Nainwal,
  ``Portalis: using competitive online interactions to support aid initiatives
  for the homeless,'' in \emph{CHI '08 extended abstracts on Human factors in
  computing systems}.\hskip 1em plus 0.5em minus 0.4em\relax New York, NY, USA:
  ACM, 2008, pp. 3873--3878. [Online]. Available:
  \url{http://portal.acm.org/citation.cfm?id=1358628.1358946}
\BIBentrySTDinterwordspacing

\bibitem{Hollis:1999:VBD:519964}
B.~S. Hollis, \emph{Visual Basic 6: Design, Specification, and Objects with
  Other}, 1st~ed.\hskip 1em plus 0.5em minus 0.4em\relax Upper Saddle River,
  NJ, USA: Prentice Hall PTR, 1999.

\bibitem{Goossens:1999:LWC:553897}
M.~Goossens, S.~P. Rahtz, R.~Moore, and R.~S. Sutor, \emph{The Latex Web
  Companion: Integrating TEX, HTML, and XML}, 1st~ed.\hskip 1em plus 0.5em
  minus 0.4em\relax Boston, MA, USA: Addison-Wesley Longman Publishing Co.,
  Inc., 1999.

\bibitem{897367}
J.~F. Buss, A.~L. Rosenberg, and J.~D. Knott, ``Vertex types in
  book-embeddings,'' Amherst, MA, USA, Tech. Rep., 1987.

\bibitem{Buss:1987:VTB:897367}
------, ``Vertex types in book-embeddings,'' Amherst, MA, USA, Tech. Rep.,
  1987.

\bibitem{Czerwinski:2008:1358628}
\emph{CHI '08: CHI '08 extended abstracts on Human factors in computing
  systems}.\hskip 1em plus 0.5em minus 0.4em\relax New York, NY, USA: ACM,
  2008, general Chair-Czerwinski, Mary and General Chair-Lund, Arnie and
  Program Chair-Tan, Desney.

\bibitem{Clarkson:1985:ACP:911891}
K.~L. Clarkson, ``Algorithms for closest-point problems (computational
  geometry),'' Ph.D. dissertation, Stanford University, Stanford, CA, USA,
  1985, aAT 8506171.

\bibitem{1984:1040142}
\emph{SIGCOMM Comput. Commun. Rev.}, vol. 13-14, no. 5-1, 1984.

\bibitem{2004:ITE:1009386.1010128}
\BIBentryALTinterwordspacing
``Ieee tcsc executive committee,'' in \emph{Proceedings of the IEEE
  International Conference on Web Services}, ser. ICWS '04.\hskip 1em plus
  0.5em minus 0.4em\relax Washington, DC, USA: IEEE Computer Society, 2004, pp.
  21--22. [Online]. Available: \url{http://dx.doi.org/10.1109/ICWS.2004.64}
\BIBentrySTDinterwordspacing

\bibitem{Mullender:1993:DS(:302430}
S.~Mullender, Ed., \emph{Distributed systems (2nd Ed.)}.\hskip 1em plus 0.5em
  minus 0.4em\relax New York, NY, USA: ACM Press/Addison-Wesley Publishing Co.,
  1993.

\bibitem{Petrie:1986:NAD:899644}
C.~J. Petrie, ``New algorithms for dependency-directed backtracking (master's
  thesis),'' Austin, TX, USA, Tech. Rep., 1986.

\bibitem{Petrie:1986:NAD:12345}
------, ``New algorithms for dependency-directed backtracking (master's
  thesis),'' Master's thesis, University of Texas at Austin, Austin, TX, USA,
  1986.

\bibitem{book-minimal}
D.~E. Knuth, \emph{Seminumerical Algorithms}.\hskip 1em plus 0.5em minus
  0.4em\relax Addison-Wesley, 1981.

\bibitem{KA:2001}
\BIBentryALTinterwordspacing
W.-C. Kong, ``The implementation of electronic commerce in smes in singapore
  (as incoll),'' in \emph{E-commerce and cultural values}.\hskip 1em plus 0.5em
  minus 0.4em\relax Hershey, PA, USA: IGI Publishing, 2001, pp. 51--74.
  [Online]. Available:
  \url{http://portal.acm.org/citation.cfm?id=887006.887010}
\BIBentrySTDinterwordspacing

\bibitem{KAGM:2001}
\BIBentryALTinterwordspacing
------, \emph{E-commerce and cultural values}.\hskip 1em plus 0.5em minus
  0.4em\relax Hershey, PA, USA: IGI Publishing, 2001, name of chapter: The
  implementation of electronic commerce in SMEs in Singapore
  (Inbook-w-chap-w-type), pp. 51--74. [Online]. Available:
  \url{http://portal.acm.org/citation.cfm?id=887006.887010}
\BIBentrySTDinterwordspacing

\bibitem{Kong:2002:IEC:887006.887010}
\BIBentryALTinterwordspacing
------, ``Chapter 9,'' in \emph{E-commerce and cultural values (Incoll-w-text
  (chap 9) 'title')}, T.~Thanasankit, Ed.\hskip 1em plus 0.5em minus
  0.4em\relax Hershey, PA, USA: IGI Publishing, 2002, pp. 51--74. [Online].
  Available: \url{http://portal.acm.org/citation.cfm?id=887006.887010}
\BIBentrySTDinterwordspacing

\bibitem{Kong:2003:IEC:887006.887011}
\BIBentryALTinterwordspacing
------, ``The implementation of electronic commerce in smes in singapore
  (incoll),'' in \emph{E-commerce and cultural values}, T.~Thanasankit,
  Ed.\hskip 1em plus 0.5em minus 0.4em\relax Hershey, PA, USA: IGI Publishing,
  2003, pp. 51--74. [Online]. Available:
  \url{http://portal.acm.org/citation.cfm?id=887006.887010}
\BIBentrySTDinterwordspacing

\bibitem{Kong:2004:IEC:123456.887010}
\BIBentryALTinterwordspacing
------, \emph{E-commerce and cultural values - (InBook-num-in-chap)}.\hskip 1em
  plus 0.5em minus 0.4em\relax Hershey, PA, USA: IGI Publishing, 2004, ch.~9,
  pp. 51--74. [Online]. Available:
  \url{http://portal.acm.org/citation.cfm?id=887006.887010}
\BIBentrySTDinterwordspacing

\bibitem{Kong:2005:IEC:887006.887010}
\BIBentryALTinterwordspacing
------, \emph{E-commerce and cultural values (Inbook-text-in-chap)}.\hskip 1em
  plus 0.5em minus 0.4em\relax Hershey, PA, USA: IGI Publishing, 2005, chapter:
  The implementation of electronic commerce in SMEs in Singapore, pp. 51--74.
  [Online]. Available:
  \url{http://portal.acm.org/citation.cfm?id=887006.887010}
\BIBentrySTDinterwordspacing

\bibitem{Kong:2006:IEC:887006.887010}
\BIBentryALTinterwordspacing
------, \emph{E-commerce and cultural values (Inbook-num chap)}.\hskip 1em plus
  0.5em minus 0.4em\relax Hershey, PA, USA: IGI Publishing, 2006, chapter (in
  type field)~22, pp. 51--74. [Online]. Available:
  \url{http://portal.acm.org/citation.cfm?id=887006.887010}
\BIBentrySTDinterwordspacing

\bibitem{SaeediMEJ10}
M.~Saeedi, M.~S. Zamani, and M.~Sedighi, ``A library-based synthesis
  methodology for reversible logic,'' \emph{Microelectron. J.}, vol.~41, no.~4,
  pp. 185--194, Apr. 2010.

\bibitem{SaeediJETC10}
M.~Saeedi, M.~S. Zamani, M.~Sedighi, and Z.~Sasanian, ``Synthesis of reversible
  circuit using cycle-based approach,'' \emph{J. Emerg. Technol. Comput.
  Syst.}, vol.~6, no.~4, Dec. 2010.

\bibitem{Kirschmer:2010:AEI:1958016.1958018}
\BIBentryALTinterwordspacing
M.~Kirschmer and J.~Voight, ``Algorithmic enumeration of ideal classes for
  quaternion orders,'' \emph{SIAM J. Comput.}, vol.~39, no.~5, pp. 1714--1747,
  Jan. 2010. [Online]. Available: \url{http://dx.doi.org/10.1137/080734467}
\BIBentrySTDinterwordspacing

\bibitem{Hoare:1972:CIN:1243380.1243382}
\BIBentryALTinterwordspacing
C.~A.~R. Hoare, ``Chapter ii: Notes on data structuring,'' in \emph{Structured
  programming (incoll)}, O.~J. Dahl, E.~W. Dijkstra, and C.~A.~R. Hoare,
  Eds.\hskip 1em plus 0.5em minus 0.4em\relax London, UK, UK: Academic Press
  Ltd., 1972, pp. 83--174. [Online]. Available:
  \url{http://portal.acm.org/citation.cfm?id=1243380.1243382}
\BIBentrySTDinterwordspacing

\bibitem{Lee:1978:TQA:800025.1198348}
\BIBentryALTinterwordspacing
J.~Lee, ``Transcript of question and answer session,'' in \emph{History of
  programming languages I (incoll)}, R.~L. Wexelblat, Ed.\hskip 1em plus 0.5em
  minus 0.4em\relax New York, NY, USA: ACM, 1981, pp. 68--71. [Online].
  Available: \url{http://doi.acm.org/10.1145/800025.1198348}
\BIBentrySTDinterwordspacing

\bibitem{Dijkstra:1979:GSC:1241515.1241518}
\BIBentryALTinterwordspacing
E.~Dijkstra, ``Go to statement considered harmful,'' in \emph{Classics in
  software engineering (incoll)}.\hskip 1em plus 0.5em minus 0.4em\relax Upper
  Saddle River, NJ, USA: Yourdon Press, 1979, pp. 27--33. [Online]. Available:
  \url{http://portal.acm.org/citation.cfm?id=1241515.1241518}
\BIBentrySTDinterwordspacing

\bibitem{Wenzel:1992:TVA:146022.146089}
\BIBentryALTinterwordspacing
E.~M. Wenzel, ``Three-dimensional virtual acoustic displays,'' in
  \emph{Multimedia interface design (incoll)}.\hskip 1em plus 0.5em minus
  0.4em\relax New York, NY, USA: ACM, 1992, pp. 257--288. [Online]. Available:
  \url{http://portal.acm.org/citation.cfm?id=146022.146089}
\BIBentrySTDinterwordspacing

\bibitem{Mumford:1987:MES:54905.54911}
\BIBentryALTinterwordspacing
E.~Mumford, ``Managerial expert systems and organizational change: some
  critical research issues,'' in \emph{Critical issues in information systems
  research (incoll)}.\hskip 1em plus 0.5em minus 0.4em\relax New York, NY, USA:
  John Wiley \& Sons, Inc., 1987, pp. 135--155. [Online]. Available:
  \url{http://portal.acm.org/citation.cfm?id=54905.54911}
\BIBentrySTDinterwordspacing

\bibitem{McCracken:1990:SSC:575315}
D.~D. McCracken and D.~G. Golden, \emph{Simplified Structured COBOL with
  Microsoft/MicroFocus COBOL}.\hskip 1em plus 0.5em minus 0.4em\relax New York,
  NY, USA: John Wiley \& Sons, Inc., 1990.

\bibitem{MR781537}
L.~H{\"o}rmander, \emph{The analysis of linear partial differential operators.
  {III}}, ser. Grundlehren der Mathematischen Wissenschaften [Fundamental
  Principles of Mathematical Sciences].\hskip 1em plus 0.5em minus 0.4em\relax
  Berlin, Germany: Springer-Verlag, 1985, vol. 275, pseudodifferential
  operators.

\bibitem{MR781536}
------, \emph{The analysis of linear partial differential operators. {IV}},
  ser. Grundlehren der Mathematischen Wissenschaften [Fundamental Principles of
  Mathematical Sciences].\hskip 1em plus 0.5em minus 0.4em\relax Berlin,
  Germany: Springer-Verlag, 1985, vol. 275, fourier integral operators.

\bibitem{Adya-01}
A.~Adya, P.~Bahl, J.~Padhye, A.Wolman, and L.~Zhou, ``A multi-radio unification
  protocol for {IEEE} 802.11 wireless networks,'' in \emph{Proceedings of the
  IEEE 1st International Conference on Broadnets Networks
  (BroadNets'04)}.\hskip 1em plus 0.5em minus 0.4em\relax Los Alamitos, CA:
  IEEE, 2004, pp. 210--217.

\bibitem{Akyildiz-01}
I.~F. Akyildiz, W.~Su, Y.~Sankarasubramaniam, and E.~Cayirci, ``Wireless sensor
  networks: A survey,'' \emph{Comm. ACM}, vol.~38, no.~4, pp. 393--422, 2002.

\bibitem{Akyildiz-02}
I.~F. Akyildiz, T.~Melodia, and K.~R. Chowdhury, ``A survey on wireless
  multimedia sensor networks,'' \emph{Computer Netw.}, vol.~51, no.~4, pp.
  921--960, 2007.

\bibitem{Bahl-02}
P.~Bahl, R.~Chancre, and J.~Dungeon, ``{SSCH}: Slotted seeded channel hopping
  for capacity improvement in {IEEE} 802.11 ad-hoc wireless networks,'' in
  \emph{Proceeding of the 10th International Conference on Mobile Computing and
  Networking (MobiCom'04)}.\hskip 1em plus 0.5em minus 0.4em\relax New York,
  NY: ACM, 2004, pp. 112--117.

\bibitem{CROSSBOW}
``{XBOW} sensor motes specifications,'' 2008, http://www.xbow.com.

\bibitem{Culler-01}
D.~Culler, D.~Estrin, and M.~Srivastava, ``Overview of sensor networks,''
  \emph{IEEE Comput.}, vol.~37, no. 8 (Special Issue on Sensor Networks), pp.
  41--49, 2004.

\bibitem{Harvard-01}
``{CodeBlue}: Sensor networks for medical care,'' 2008,
  http://www.eecs.harvard.edu/mdw/ proj/codeblue/.

\bibitem{Natarajan-01}
A.~Natarajan, M.~Motani, B.~de~Silva, K.~Yap, and K.~C. Chua, ``Investigating
  network architectures for body sensor networks,'' in \emph{Network
  Architectures}, G.~Whitcomb and P.~Neece, Eds.\hskip 1em plus 0.5em minus
  0.4em\relax Dayton, OH: Keleuven Press, 2007, pp. 322--328.

\bibitem{Tzamaloukas-01}
A.~Tzamaloukas and J.~J. Garcia-Luna-Aceves, ``Channel-hopping multiple
  access,'' Department of Computer Science, University of California, Berkeley,
  CA, Tech. Rep. I-CA2301, 2000.

\bibitem{Zhou-06}
G.~Zhou, J.~Lu, C.-Y. Wan, M.~D. Yarvis, and J.~A. Stankovic, \emph{Body Sensor
  Networks}.\hskip 1em plus 0.5em minus 0.4em\relax Cambridge, MA: MIT Press,
  2008.

\bibitem{ko94}
J.~Kornerup, ``Mapping powerlists onto hypercubes,'' Master's thesis, The
  University of Texas at Austin, 1994, (In preparation).

\bibitem{gerndt:89}
M.~Gerndt, ``Automatic parallelization for distributed-memory multiprocessing
  systems,'' Ph.D. dissertation, University of Bonn, Bonn, Germany, Dec. 1989.

\bibitem{6:1:1}
J.~E. {Archer, Jr.}, R.~Conway, and F.~B. Schneider, ``User recovery and
  reversal in interactive systems,'' \emph{ACM Trans. Program. Lang. Syst.},
  vol.~6, no.~1, pp. 1--19, Jan. 1984.

\bibitem{7:1:137}
D.~D. Dunlop and V.~R. Basili, ``Generalizing specifications for uniformly
  implemented loops,'' \emph{ACM Trans. Program. Lang. Syst.}, vol.~7, no.~1,
  pp. 137--158, Jan. 1985.

\bibitem{7:2:183}
J.~Heering and P.~Klint, ``Towards monolingual programming environments,''
  \emph{ACM Trans. Program. Lang. Syst.}, vol.~7, no.~2, pp. 183--213, Apr.
  1985.

\bibitem{knuth:texbook}
D.~E. Knuth, \emph{The {\TeX{}book}}.\hskip 1em plus 0.5em minus 0.4em\relax
  Reading, MA.: Addison-Wesley, 1984.

\bibitem{6:3:380}
E.~Korach, D.~Rotem, and N.~Santoro, ``Distributed algorithms for finding
  centers and medians in networks,'' \emph{ACM Trans. Program. Lang. Syst.},
  vol.~6, no.~3, pp. 380--401, Jul. 1984.

\bibitem{lamport:latex}
L.~Lamport, \emph{\it {\LaTeX}: A Document Preparation System}.\hskip 1em plus
  0.5em minus 0.4em\relax Reading, MA.: Addison-Wesley, 1986.

\bibitem{7:3:359}
F.~Nielson, ``Program transformations in a denotational setting,'' \emph{ACM
  Trans. Program. Lang. Syst.}, vol.~7, no.~3, pp. 359--379, Jul. 1985.

\bibitem{test}
D.~E. Knuth, \emph{Seminumerical Algorithms}, 2nd~ed., ser. The Art of Computer
  Programming.\hskip 1em plus 0.5em minus 0.4em\relax Reading, MA:
  Addison-Wesley, 10~Jan. 1981, vol.~2.

\bibitem{reid:scribe}
B.~K. Reid, ``A high-level approach to computer document formatting,'' in
  \emph{Proceedings of the 7th Annual Symposium on Principles of Programming
  Languages}.\hskip 1em plus 0.5em minus 0.4em\relax New York: ACM, Jan. 1980,
  pp. 24--31.

\bibitem{Zhou:2010:MMS:1721695.1721705}
\BIBentryALTinterwordspacing
G.~Zhou, Y.~Wu, T.~Yan, T.~He, C.~Huang, J.~A. Stankovic, and T.~F. Abdelzaher,
  ``A multifrequency mac specially designed for wireless sensor network
  applications,'' \emph{ACM Trans. Embed. Comput. Syst.}, vol.~9, no.~4, pp.
  39:1--39:41, April 2010. [Online]. Available:
  \url{http://doi.acm.org/10.1145/1721695.1721705}
\BIBentrySTDinterwordspacing

\bibitem{TUGInstmem}
\BIBentryALTinterwordspacing
(2017) Institutional members of the {\TeX} users group. [Online]. Available:
  \url{http://wwtug.org/instmem.html}
\BIBentrySTDinterwordspacing

\bibitem{CTANacmart}
\BIBentryALTinterwordspacing
B.~Veytsman. acmart---{C}lass for typesetting publications of {ACM}. [Online].
  Available: \url{http://www.ctan.org/pkg/acmart}
\BIBentrySTDinterwordspacing

\bibitem{bowman:reasoning}
M.~Bowman, S.~K. Debray, and L.~L. Peterson, ``Reasoning about naming
  systems,'' \emph{ACM Trans. Program. Lang. Syst.}, vol.~15, no.~5, pp.
  795--825, November 1993.

\bibitem{braams:babel}
J.~Braams, ``Babel, a multilingual style-option system for use with latex's
  standard document styles,'' \emph{TUGboat}, vol.~12, no.~2, pp. 291--301,
  June 1991.

\bibitem{clark:pct}
M.~Clark, ``Post congress tristesse,'' in \emph{TeX90 Conference
  Proceedings}.\hskip 1em plus 0.5em minus 0.4em\relax TeX Users Group, March
  1991, pp. 84--89.

\bibitem{herlihy:methodology}
M.~Herlihy, ``A methodology for implementing highly concurrent data objects,''
  \emph{ACM Trans. Program. Lang. Syst.}, vol.~15, no.~5, pp. 745--770,
  November 1993.

\bibitem{salas:calculus}
S.~Salas and E.~Hille, \emph{Calculus: One and Several Variable}.\hskip 1em
  plus 0.5em minus 0.4em\relax New York: John Wiley and Sons, 1978.

\bibitem{Fear05}
S.~Fear, \emph{Publication quality tables in {\LaTeX}}, April 2005,
  \url{http://www.ctan.org/pkg/booktabs}.

\bibitem{Amsthm15}
\emph{Using the amsthm Package}, American Mathematical Society, April 2015,
  \url{http://www.ctan.org/pkg/amsthm}.

\bibitem{R}
\BIBentryALTinterwordspacing
{R Core Team}, ``R: A language and environment for statistical computing,'' R
  Foundation for Statistical Computing, Vienna, Austria, 2019. [Online].
  Available: \url{https://www.R-project.org/}
\BIBentrySTDinterwordspacing

\bibitem{UMassCitations}
\BIBentryALTinterwordspacing
S.~Anzaroot and A.~McCallum, ``{UMass} citation field extraction dataset,''
  2013. [Online]. Available:
  \url{http://www.iesl.cs.umass.edu/data/data-umasscitationfield}
\BIBentrySTDinterwordspacing

\bibitem{vaswani2017attention}
\BIBentryALTinterwordspacing
A.~Vaswani, N.~Shazeer, N.~Parmar, J.~Uszkoreit, L.~Jones, A.~N. Gomez, L.~u.
  Kaiser, and I.~Polosukhin, ``Attention is all you need,'' in \emph{Advances
  in Neural Information Processing Systems 30}, I.~Guyon, U.~V. Luxburg,
  S.~Bengio, H.~Wallach, R.~Fergus, S.~Vishwanathan, and R.~Garnett, Eds.\hskip
  1em plus 0.5em minus 0.4em\relax Curran Associates, Inc., 2017, pp.
  5998--6008. [Online]. Available:
  \url{http://papers.nips.cc/paper/7181-attention-is-all-you-need.pdf}
\BIBentrySTDinterwordspacing

\bibitem{DBLP:journals/corr/abs-2003-11080}
\BIBentryALTinterwordspacing
J.~Hu, S.~Ruder, A.~Siddhant, G.~Neubig, O.~Firat, and M.~Johnson, ``{XTREME:}
  {A} massively multilingual multi-task benchmark for evaluating cross-lingual
  generalization,'' \emph{CoRR}, vol. abs/2003.11080, 2020. [Online].
  Available: \url{https://arxiv.org/abs/2003.11080}
\BIBentrySTDinterwordspacing

\bibitem{10.5555/1177220}
A.~V. Aho, M.~S. Lam, R.~Sethi, and J.~D. Ullman, \emph{Compilers: Principles,
  Techniques, and Tools (2nd Edition)}.\hskip 1em plus 0.5em minus 0.4em\relax
  USA: Addison-Wesley Longman Publishing Co., Inc., 2006.

\bibitem{devlin-etal-2019-bert}
\BIBentryALTinterwordspacing
J.~Devlin, M.-W. Chang, K.~Lee, and K.~Toutanova, ``{BERT}: Pre-training of
  deep bidirectional transformers for language understanding,'' in
  \emph{Proceedings of the 2019 Conference of the North {A}merican Chapter of
  the Association for Computational Linguistics: Human Language Technologies,
  Volume 1 (Long and Short Papers)}.\hskip 1em plus 0.5em minus 0.4em\relax
  Minneapolis, Minnesota: Association for Computational Linguistics, Jun. 2019,
  pp. 4171--4186. [Online]. Available:
  \url{https://www.aclweb.org/anthology/N19-1423}
\BIBentrySTDinterwordspacing

\bibitem{aho1986compilers}
A.~V. Aho, R.~Sethi, and J.~D. Ullman, ``Compilers, principles, techniques,''
  \emph{Addison wesley}, vol.~7, no.~8, p.~9, 1986.

\bibitem{darki2021disco}
A.~Darki, M.~Faloutsos, N.~Abu-Ghazaleh, and M.~Sridharan, ``Disco: Combining
  disassemblers for improved performance,'' 2021.

\bibitem{pei2020xda}
K.~Pei, J.~Guan, D.~Williams-King, J.~Yang, and S.~Jana, ``Xda: Accurate,
  robust disassembly with transfer learning,'' \emph{arXiv preprint
  arXiv:2010.00770}, 2020.

\bibitem{2015arXiv151105493L}
Y.~{Li}, D.~{Tarlow}, M.~{Brockschmidt}, and R.~{Zemel}, ``{Gated Graph
  Sequence Neural Networks},'' \emph{arXiv e-prints}, p. arXiv:1511.05493, Nov.
  2015.

\bibitem{tensorflow2015-whitepaper}
\BIBentryALTinterwordspacing
M.~Abadi, A.~Agarwal, P.~Barham, E.~Brevdo, Z.~Chen, C.~Citro, G.~S. Corrado,
  A.~Davis, J.~Dean, M.~Devin, S.~Ghemawat, I.~Goodfellow, A.~Harp, G.~Irving,
  M.~Isard, Y.~Jia, R.~Jozefowicz, L.~Kaiser, M.~Kudlur, J.~Levenberg,
  D.~Man\'{e}, R.~Monga, S.~Moore, D.~Murray, C.~Olah, M.~Schuster, J.~Shlens,
  B.~Steiner, I.~Sutskever, K.~Talwar, P.~Tucker, V.~Vanhoucke, V.~Vasudevan,
  F.~Vi\'{e}gas, O.~Vinyals, P.~Warden, M.~Wattenberg, M.~Wicke, Y.~Yu, and
  X.~Zheng, ``{TensorFlow}: Large-scale machine learning on heterogeneous
  systems,'' 2015, software available from tensorflow.org. [Online]. Available:
  \url{http://tensorflow.org/}
\BIBentrySTDinterwordspacing

\bibitem{residual}
\BIBentryALTinterwordspacing
K.~He, X.~Zhang, S.~Ren, and J.~Sun, ``Deep residual learning for image
  recognition,'' \emph{CoRR}, vol. abs/1512.03385, 2015. [Online]. Available:
  \url{http://arxiv.org/abs/1512.03385}
\BIBentrySTDinterwordspacing

\bibitem{kullback1951information}
S.~Kullback and R.~A. Leibler, ``On information and sufficiency,'' \emph{The
  annals of mathematical statistics}, vol.~22, no.~1, pp. 79--86, 1951.

\bibitem{li2015gated}
Y.~Li, D.~Tarlow, M.~Brockschmidt, and R.~Zemel, ``Gated graph sequence neural
  networks,'' \emph{arXiv preprint arXiv:1511.05493}, 2015.

\bibitem{li2019graph2seq}
W.~Li, X.~Zhang, Y.~Wang, Z.~Yan, and R.~Peng, ``Graph2seq: Fusion embedding
  learning for knowledge graph completion,'' \emph{IEEE Access}, vol.~7, pp.
  157\,960--157\,971, 2019.

\end{thebibliography}
\end{document}